%% file: acl2021.tex
\documentclass[11pt,a4paper]{article}
\usepackage[hyperref]{acl2021}
\usepackage{times}
\usepackage{latexsym}
\usepackage{booktabs}
\usepackage{amstext}
\usepackage{amsmath}
\usepackage{multirow}
\usepackage{color}
\usepackage{array}
\usepackage{graphicx}
\usepackage{subfigure}
\newcommand\name{KACC}

\definecolor{myLightCyan}{rgb}{0.831,0.875,0.921}
\usepackage{tabularx}
\usepackage{xcolor}
\usepackage{colortbl}
\usepackage{tikz-cd}
\usetikzlibrary{shapes.misc, positioning}
\usetikzlibrary{shapes.geometric, arrows,calc}
\tikzstyle{roundrec} = [rectangle, rounded corners, minimum width=1.4cm, minimum height=0.4cm,text height=0.23cm,
    text depth=0.11cm, text centered, draw=black,]
\tikzstyle{arrow} = [->,>=stealth]
\usepackage{microtype}

\newcommand*{\Bhline}[0]{\noalign{\global\setlength{\arrayrulewidth}{0.9pt}}
\hline\noalign{\global\setlength{\arrayrulewidth}{0.4pt}}}

\aclfinalcopy % Uncomment this line for the final submission
\def\aclpaperid{1305} %  Enter the acl Paper ID here

\title{KACC: A Multi-task Benchmark for Knowledge Abstraction, Concretization and Completion}

\author{Jie Zhou$^{1,2*}$, Shengding Hu$^{1,2*}$, Xin Lv$^{1,2}$, Cheng Yang$^{4}$, \\  \textbf{Zhiyuan Liu$^{1,2,3\dagger}$, Wei Xu$^{5}$, Jie Jiang$^{5}$, Juanzi Li$^{1,2}$, Maosong Sun$^{1,2,3}$}\\
\textsuperscript{1}Department of Computer Science and Technology \quad
\textsuperscript{2}Institute for Artificial Intelligence \\
\textsuperscript{3}State Key Lab on Intelligent Technology and Systems, Tsinghua University, Beijing, China\\
\textsuperscript{4}School of Computer Science, Beijing University of Posts and Telecommunications, China\\
\textsuperscript{5}Tencent Marketing Solution, Tencent, Shenzhen, China \\
\texttt{\{zhoujie18, hsd20, lv-x18\}@mails.tsinghua.edu.cn} \\
}

\date{}

\begin{document}
\maketitle
\begin{abstract}
A comprehensive knowledge graph (KG) contains an instance-level entity graph and an ontology-level concept graph. 
The two-view KG provides a testbed for models to ``simulate'' human's abilities on knowledge abstraction, concretization, and completion (KACC),
which are crucial for human to recognize the world and manage learned knowledge.
Existing studies mainly focus on partial aspects of KACC. In order to promote thorough analyses for KACC abilities of models, we propose a unified KG benchmark by improving existing benchmarks in terms of dataset scale, task coverage, and difficulty.
Specifically, we collect new datasets that contain larger concept graphs, abundant cross-view links as well as dense entity graphs. Based on the datasets, we propose novel tasks such as multi-hop knowledge abstraction (MKA), multi-hop knowledge concretization (MKC) and then design a comprehensive benchmark. 
For MKA and MKC tasks, we further annotate multi-hop hierarchical triples as harder samples.
The experimental results of existing methods demonstrate the challenges of our benchmark. The resource is available at \url{https://github.com/thunlp/KACC}.

\end{abstract}

\section{Introduction}
{\let\thefootnote\relax\footnotetext{$*$ indicates equal contribution}
\let\thefootnote\relax\footnotetext{$^\dagger$ Corresponding author: Z.Liu(liuzy@tsinghua.edu.cn)}}

\begin{figure}[t]
    \centering
    \includegraphics[width=0.9\linewidth]{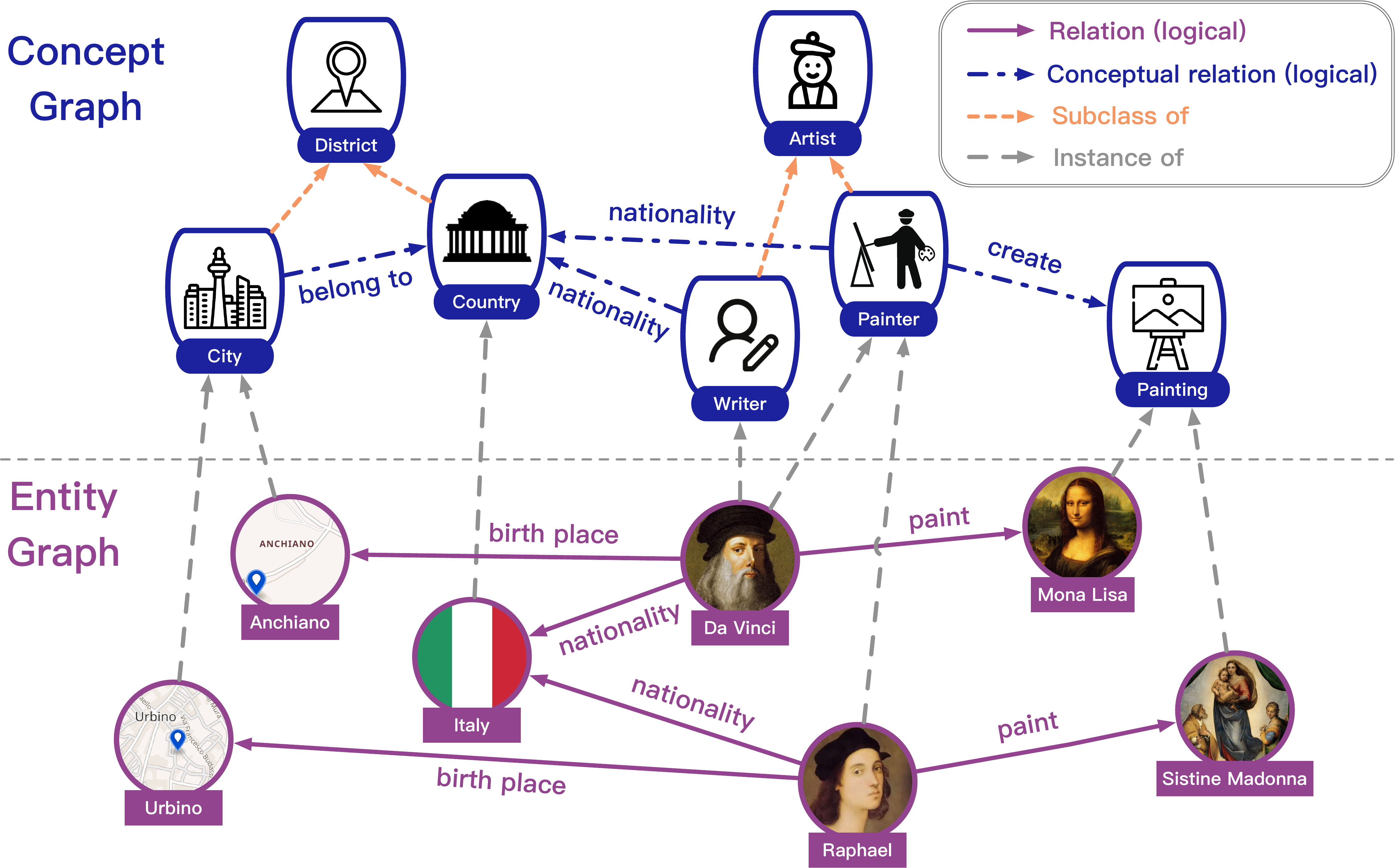}
    \vspace{-0.5em}
    \caption{An example of the entity-concept KG.}
    \label{fig:kg}
    \vspace{-1.5em}
\end{figure}

Large-scale knowledge graphs (KGs) like Wikidata~\cite{vrandevcic2014wikidata}, DBpedia~\cite{lehmann2015dbpedia}, and YAGO~\cite{mahdisoltani2013yago3} usually contain two subgraphs: an instance-level \textbf{entity graph} and an ontology-level \textbf{concept graph}. The entity graph (a.k.a. the entity view) is composed of entities and relations. It describes factual knowledge such as {(\emph{Da Vinci}, \texttt{paint}, \emph{Mona Lisa})}. The concept graph (a.k.a. the concept view) contains concepts and conceptual relations. It provides abstract and commonsense knowledge like {(\emph{painter}, \texttt{create}, \emph{painting})}. In this paper, we name this kind of two-view KG as the entity-concept KG (EC-KG). 
{In a EC-KG, the relations can be grouped into three categories.  The ``\texttt{subclassOf}'' relation forms hierarchical concept structures via triples like (\textit{painter}, \texttt{subclassOf}, \textit{artist}). The ``\texttt{instanceOf}'' relation groups entities into concepts, such as (\textit{Da Vinci}, \texttt{instanceOf}, \textit{painter}). These two relations are important for testing models' abilities on knowledge abstraction and concretization. Other relations are \emph{logical} relations for testing models' abilities on knowledge completion. } An example of the EC-KG is shown in Figure~\ref{fig:kg}. 

During the last decade, there are massive works focusing on learning representations for KGs such as TransE~\cite{bordes2013translating}, DistMult~\cite{yang2015embedding}, ComplEx~\cite{trouillon2016complex}, and TuckER~\cite{balavzevic2019tucker}. Though they have achieved promising results on knowledge graph completion, most of them focus
on a single graph, especially the entity graph.
% , and pay less attention to the other two ability of the models.

Beyond modeling a single graph of KGs, recent studies demonstrate that jointly modeling the two graphs in the EC-KG can improve the understanding of each one~\cite{xie2016representation, moon2017learning, lv2018differentiating, hao2019universal}. 
They also propose several tasks  on the EC-KG, such as link prediction and entity typing. These tasks focus on partial aspects of knowledge abstraction, concretization, and completion, which are essential abilities for humans to recognize the world and acquire knowledge.
For example, in entity typing, a model may link the entity ``\emph{Da Vinci}'' to the concept ``\emph{painter}'' 
% according to its entity graph triples,
which reflects the model's abstraction ability. 
However, 
% most works merely focus on partial tasks of KACC and 
little work has been devoted to unified benchmarking and studies on \name.

In this paper, we present a comprehensive benchmark for KACC by improving existing benchmarks in 
% terms of  %%to save a line
dataset scale, task coverage, and difficulty.

\textbf{Dataset scale.} We have examined 
% existin
the EC-KGs proposed by previous works such as  ~\citet{hao2019universal}.
% based on DBpedia and YAGO. 
Due to the data distribution, 
the concept graphs are 
% relatively
small compared to the entity graphs. Furthermore, the cross-links between the two graphs are also sparse (refer to~Section \ref{sec:data_analysis}). These may limit the knowledge transfer between the two graphs.
% and cannot provide sufficient information for knowledge abstraction and concretization.
To tackle these problems, we construct several different-scale datasets based on Wikidata~\cite{vrandevcic2014wikidata} with careful filtering, annotation and refinement. As Wikidata contains more fine-grained concepts, our datasets have large concept graphs, abundant cross-view links, 
as well as dense entity graphs.

\textbf{Task coverage.} Most previous works focus on partial tasks of KACC.
% such as link prediction and entity typing. In our benchmark,
% We summarize existing tasks and further propose two new tasks.
In our benchmark, we define the tasks more comprehensively and categorize these tasks into three classes: knowledge abstraction,  concretization, and  completion. 

\textbf{Difficulty.} We propose two new tasks, including multi-hop knowledge abstraction and multi-hop knowledge concretization,
% . 
% These tasks are meaningful and important as they require models to predict for multi-hop ``\texttt{instanceOf}'' and ``\texttt{subclassOf}'' triples, which do not exist in KGs but can be inferred via relation transitivity. 
{which require models to predict multi-hop ``\texttt{instanceOf}'' and ``\texttt{subclassOf}'' triples that do not exist in KGs but can be inferred via relation transitivity. These tasks are meaningful and important since correctly modeling these triples is necessary for models to truly understand the concept hierarchy.} 
% Abilities of multi-hop abstraction and concretization are also crucial for multi-hop human cognition.
% They use multi-hop ``\texttt{instanceOf}'' and ``\texttt{subclassOf}'' triples which are inferred via relation transitivity. 
To ensure the quality of these tasks, we annotate corresponding multi-hop datasets.
% and the experimental results have shown these triples are more difficult for existing models.
{Our experiments show that these tasks are still challenging for existing models.}

Based on our benchmark, we conduct extensive experiments for existing baselines and provide thorough analyses. The experiments show 
that {while the methods specifically designed for modeling hierarchies perform better than general KGE models on abstraction and concretization tasks, they are not competitive to some general KGE models on logical relations. Moreover, all methods have drastic performance degradation on multi-hop tasks, and the knowledge transfer between the entity graph and the concept graph is still obscure.} 
% existing models can only perform well on partial tasks of KACC and there lacks a model that can jointly handle all proposed tasks. 
% logical and hierarchical relations well. 
Finally, we present useful insights for future model design.
% The experiments demonstrate the challenges of KACC and provide useful insights for the future model design. % Our codes and datasets will be released.

\section{Related Work}
% In this section, we introduce relevant datasets and knowledge embedding methods on the EC-KG.
\subsection{Knowledge Graph Datasets}
Existing datasets for knowledge graph completion are usually subgraphs of large-scale KGs, such as 
% FB15K~\cite{bordes2013translating}, FB15K-237~\cite{toutanova2015representing}, WN18~\cite{bordes2013translating}, WN18RR~\cite{dettmers2018convolutional}, and CoDEx~\cite{safavi2020codex}. 
FB15K, FB15K-237, WN18, WN18RR and CoDEx~\cite{bordes2013translating, toutanova2015representing,dettmers2018convolutional,safavi2020codex}.
% FB15K-237~\cite{toutanova2015representing}, WN18~\cite{bordes2013translating}, WN18RR~\cite{dettmers2018convolutional}, and CoDEx~\cite{safavi2020codex}. 
These datasets are all single-view KGs, in which FB15K, FB15K-237, and CoDEx focus on the entity view while WN18 and WN18RR can be regarded as concept view KGs.
% \hsd{For example, FB15K, FB15K-237 and CoDEx  contains few concepts while WN18, WN18RR provide concept graphs in the absent of complex relational triples between real world entities.} 
% For example, FB15K, FB15K-237 and CoDEx focus on the entity view while WN18 and WN18RR can be regarded as concept view KGs. 
Several  datasets try to link the two views in different ways. Firstly, some datasets provide additional type information to the entity graph, such as FB15K+, FB15K-ET and YAGO43K-ET~\cite{xie2016representation,moon2017learning}.
% incorporate entity types into single-view KGs.
Secondly, some datasets provide concept hierarchies for the entity graph, such as Probase~\cite{wu2012probase} and YAGO39K~\cite{lv2018differentiating}. However, they do not provide the full concept graphs with logical relations. Thirdly, some datasets provide the full concept graphs~\cite{hao2019universal}, but both the scale and the depth of the concept hierarchy are limited. 
For example, the entity numbers of DB111K-174~\cite{hao2019universal} and our  dataset KACC-M are similar, but KACC-M has 38 times more concepts than DB111K-174 (see Table~\ref{tab:dataset}).
% For example, in the two datasets proposed by~\citet{hao2019universal},
% the ratios of the number of concepts to the number of entities are 3.5\% and 0.15\%, respectively, while the ratio is 21\% and in our constructed datasets.

% Some datasets like FB15K+~\cite{xie2016representation}, FB15K-ET, and YAGO43K-ET~\cite{moon2017learning} incorporate type information of entities into single-view KGs. However, they only provide entity-concept links and lack relations between concepts.
% Other datasets like Probase~\cite{wu2012probase}, YAGO39K~\cite{lv2018differentiating} further provide concept hierarchies in a tree structure. These datasets do not provide the graph structure of concepts due to the lack of meta-relations.
% ~\citet{hao2019universal} propose two datasets YAGO26K-906 and DB111K-174 which provide concept graphs with meta-relations. However, the concept graphs in YAGO26K-906 and DB111K-174 are limited in both scale and the depth of hierarchy, which can not support more comprehensive studies of KACC.

\iffalse
Besides these KGs for factual knowledge, several multi-view KGs for domain knowledge are also proposed. Recent published AliCoCo~\cite{luo2020alicoco} and the Attention Ontology~\cite{liu2020giant} are KGs constructed for e-commerce recommendation. In contrast, the datasets in our benchmark are  extracted from Wikidata, which contain a wide range of domains.
\fi

\subsection{Knowledge Embedding Methods}
Existing knowledge embedding (KE) methods can be categorized as translation models~\cite{bordes2013translating, wang2014knowledge, lin2015learning, ji2015knowledge, sun2018rotate}, tensor factorization based models~\cite{yang2015embedding, nickel2016holographic, trouillon2016complex, balavzevic2019tucker}, and neural models~\cite{socher2013reasoning, dettmers2018convolutional, nguyen2018novel}. These methods are typically designed for 
% learning knowledge embeddings on
single-view KGs.  Although they can be directly applied to EC-KGs by ignoring different characteristics between entity graphs and concept graphs, they cannot take full advantage of the information in EC-KGs.  

Several works ~\cite{krompass2015type, xie2016representation,ma2017transt, moon2017learning} incorporate the type information into KE methods to help the completion of entity graphs. ETE~\cite{moon2017learning} further conducts entity typing, which can be seen as a simplified version of our knowledge abstraction task.  Though types of entities can be seen as concepts, they omit the concept hierarchy and interactions (conceptual relations) between concepts. 

\iffalse
For the Knowledge Abstraction task alone, some works~\cite{tifrea2018poincar,athiwaratkun2018hierarchical,sala2018representation,nickel2018learning} prove the effectiveness of  Gaussian embedding and hyperbolic embedding for encoding the hierarchical structure. However they can not capture logical relations in the EC-KG at the same time. 
\fi

To jointly model the whole EC-KG, 
% TransC~\cite{lv2018differentiating}  uses a simple model for predicting ``\texttt{subclassOf}'' hierarchies and inherits the ability of KGE methods for modeling entity graphs. 
TransC~\cite{lv2018differentiating} adopts TransE as the entity graph model and models concepts as spheres that enclosing points of entities. However,
% It inherits the ability of KGE methods for modeling entity graphs, 
it is not flexible enough to model logical relations between concepts. AttH~\cite{chami2020low} further combines hyperbolic embedding methods with KGE methods to simultaneously embed hierarchical and logical relations.
% However our benchmark contains more challenging dataset, which Our Their tasks can be seen as "Knowledge Abstraction" part of our benchmark. And our benchmark also contains complex facteral knowledge and entities and concept.
% Beyond the task of entity graph completion, ETE~\cite{moon2017learning} tries to predict concepts for entities and TransC~\cite{lv2018differentiating} further predicts for ``\texttt{subclassOf}'' relations. 
JOIE~\cite{hao2019universal} uses different training paradigms for training the entity graph, the concept graph, and the cross-view links. 
% They first work that attempts to embed the EC-KG. It adopts cross-view and intra-view models to learn different facets of the EC-KG and achieves promising results. 
% \hsd{Their results benefit from embedding entities and concepts into different spaces. In our experiment, we show that some recently proposed models can achieve promising results when embedding entities and concepts into the same spaces. }
It also defines several meaningful tasks on the EC-KG. In this paper, we extend these tasks with several newly proposed tasks, then we categorize and formalize these tasks into a unified benchmark. We also test several KE methods as mentioned above using our benchmarks and analyze their advantages and deficiencies in terms of handling these tasks.
% There are also several models from other research branch such as multi-graph KG embedding methods like MTransE~\cite{chen2016multilingual}. However, these methods are originally designed for knowledge graph alignment and they need similar structures of two KGs. 
\vspace{-0.3em}
\section{Benchmark}
\vspace{-0.4em}
In this section, we propose the {\name} benchmark with three tasks: knowledge abstraction, knowledge concretization, and knowledge completion.
\vspace{-1em}
\subsection{Formalizations}
\label{sec:formalization}
We first give formalizations of the EC-KG, then we introduce multi-hop 
% hierarchical 
triples used in later tasks.

\textbf{Formalizations of EC-KG.} 
% A knowledge graph triple $(h, r, t)$, where $h, r, t$ denotes the head, relation, and tail of the triple, respectively.
A EC-KG is a comprehensive KG, which contains two subgraphs and the cross-view links. The entity graph $\mathcal{G}_E = \{ \mathcal{E}_E, \mathcal{R}_E, \mathcal{T}_E \}$ is composed of the entity set $\mathcal{E}_E$, relation set $\mathcal{R}_E$, and corresponding triple set $\mathcal{T}_E=\{(h^E, r^E, t^E)~|~h^E, t^E\in \mathcal{E}_E, r^E\in \mathcal{R}_E\}$, where $h, r, t$ represent head, relation, and tail of a triple, respectively. 
% We use $(h^E, r^E, t^E)$ to denote triples in the entity graph and we have $\mathcal{T}_E = \{(h^E, r^E, t^E) \; | \; h^E, t^E \in \mathcal{E}_E, r^E \in \mathcal{R}_E\}$.
The concept graph $\mathcal{G}_C = \{ \mathcal{E}_C, \mathcal{R}_C, \mathcal{T}_C \}$ contains the concept set $\mathcal{E}_C$, conceptual relation set $\mathcal{R}_C$, and triple set $\mathcal{T}_C$.
%$\mathcal{T}_C = \{(h^C, r^C, t^C) \; | \; h^C, t^C \in \mathcal{E}_C, r^C \in \mathcal{R}_C\}$
In our settings, $\mathcal{E}_E$ and $\mathcal{E}_C$ are disjoint sets, while $\mathcal{R}_E$ and $\mathcal{R}_C$ may contain some relations in common (see Section~\ref{sec:data_analysis}). The cross-view links $\mathcal{T}_S = \{(h^S,r_{\texttt{ins}},t^S)\}$ connects the two subgraphs, where $h^S\in \mathcal{E}_E, t^S\in \mathcal{E}_C$, and $r_{\texttt{ins}}$ is the ``\texttt{instanceOf}'' relation.
% have head entity set .  contains triples from entities to concepts.
% $\mathcal{T}_{\texttt{ins}} = \mathcal{T}_S \cup \mathcal{T}_{C(\texttt{ins})}$ and $\mathcal{T}_{\texttt{sub}} = \mathcal{T}_{C(\texttt{sub})}$ are the full sets of ``\texttt{instanceOf}'' and ``\texttt{subclassOf}'' triples.
% The EC-KG $\mathcal{G}$ is a combination of the above three parts, thus we have:
Therefore, the EC-KG is
$\mathcal{G} = \{ \mathcal{E}_{EC}, \mathcal{R}_{EC}, \mathcal{T}_{EC} \}$, where $\mathcal{E}_{EC} = \mathcal{E}_E \cup \mathcal{E}_C$, $\mathcal{R}_{EC} = \mathcal{R}_E \cup \mathcal{R}_C \cup \{r_{\texttt{ins}}\}$, and $\mathcal{T}_{EC} = \mathcal{T}_E \cup \mathcal{T}_C \cup \mathcal{T}_S$.

{There are two special relations ``\texttt{instanceOf}'' and ``\texttt{subclassOf}'' that are crucial for knowledge abstraction and concretization. We use ``\texttt{ins}'' and ``\texttt{sub}'' to denote them in the rest of our paper, respectively. The corresponding triples are $\mathcal{T}_{\texttt{ins}} = \mathcal{T}_S$ and $\mathcal{T}_{\texttt{sub}}\subset\mathcal{T}_C $. Other relations are \emph{logical} relations. Their corresponding triples in concept graphs are $\mathcal{T}_{C(\texttt{logic})}= \mathcal{T}_{C} \backslash \mathcal{T}_{{\texttt{sub}}}$ and logical triples in the entity graphs are $\mathcal{T}_{E(\texttt{logic})}= \mathcal{T}_E$. }
% As there are also ``\texttt{instanceOf}'' triples in the concept graph (see Section~\ref{sec:data_analysis}), we use $\mathcal{T}_{C(\texttt{ins})}$ to denote them. 
% We use $r^C_{\texttt{sub}}$ to denote the ``\texttt{subclassOf}'' relation and $\mathcal{T}_{\texttt{sub}} = \mathcal{T}_{C(\texttt{sub})}$ denotes a subset of $\mathcal{T}_C$ with all ``\texttt{subclassOf}'' triples. 
% We use $\mathcal{T}_{C(\texttt{sub})}$ to denote a subset of $\mathcal{T}_C$ with all ``\texttt{subclassOf}'' triples. 
% $\mathcal{T}_{C(\texttt{logic})}$ denotes logical triples in the concept graph apart from hierarchical relations, that is $\mathcal{T}_{C(\texttt{logic})} = \mathcal{T}_{C} \backslash \{ \mathcal{T}_{C{(\texttt{ins})}} \cup \mathcal{T}_{C{(\texttt{sub})}} \}$.

\textbf{Multi-hop Triples.}
Hierarchical relations like ``\texttt{ins}'' and ``\texttt{sub}'' should preserve the multi-hop transitivity, which can be explained by two rules:
% \begin{equation}
% \begin{split}
\vspace{-1.0em}

\begin{equation}
\small
\begin{split}
{}(e_{\phantom{}},\texttt{ins},c_1)\wedge & {}(c_1, \texttt{sub}, c_2) \wedge ... \wedge \\
 &{}(c_{N-1},\texttt{sub}, c_{N}) \Rightarrow {}(e_{\phantom{}}, \texttt{ins},c_N) ,
 \end{split}
  \label{equ:1}
 \end{equation}
 \vspace{-2.0em}
 \begin{equation}
 \small
 \begin{split}
{}(c_0,\texttt{sub}, c_1) \wedge & {}(c_1,\texttt{sub}, c_2) \wedge ... \wedge \\
 & {}(c_{N-1}, \texttt{sub},c_{N}) \Rightarrow {}(c_0,\texttt{sub}, c_N), 
 \end{split}
 \label{equ:2}
 \end{equation}
% \normalsize
% \end{split0-
% \end{equation}
in which $\{c_i|i\geq 1\}$ are defined as the \emph{high-level concept} for $e$ and $c_0$. These two rules indicate that an entity/concept always belongs to its high-level concepts.
% With these rules, high-order hierarchical triples like ($e$, \texttt{ins}, $c_N$) and ($c_1$, \texttt{sub}, $c_N$) can be collected from the train data.
With these rules, we can collect multi-hop hierarchical triples like ($e$, \texttt{ins}, $c_N$) and ($c_0$, \texttt{sub}, $c_N$) from the train data and use them as harder samples for knowledge abstraction and concretization testing. 
Corresponding datasets of multi-hop hierarchical triples are denoted as $\mathcal{T}_{\texttt{M-Ins}}$ and $\mathcal{T}_{\texttt{M-Sub}}$.
% With these rules, we can collect multi-hop hierarchical triples like ($e$, \texttt{ins}, $c_N$) and ($c_1$, \texttt{sub}, $c_N$) from the train data and use them as harder samples for knowledge abstraction and concretization. 
% As multi-hop triples extracted from the datasets do not always hold true, we further annotate these triples and the details of the annotation process are shown in Section~\ref{sec:annotation}. 

\subsection{Knowledge Abstraction}
This task contains tail prediction tasks for one-hop and multi-hop ``\texttt{ins}'' and ``\texttt{sub}'' triples. We use KA (knowledge abstraction) and MKA (multi-hop knowledge abstraction) to denote the tasks.

\textbf{KA-Ins / KA-Sub:} KA-Ins and KA-Sub are tail prediction tasks for ``\texttt{ins}'' triples 
% in $\mathcal{T}_{\texttt{ins}}$ 
and ``\texttt{sub}'' triples
% in $\mathcal{T}_{\texttt{sub}}$
respectively. These are all  triples in the original datasets and these tasks reflect the direct knowledge abstraction ability of models.

\textbf{MKA-Ins / MKA-Sub:} MKA-Ins and MKA-Sub are tail prediction tasks for multi-hop hierarchical triples $\mathcal{T}_{\texttt{M-Ins}}$ and $\mathcal{T}_{\texttt{M-Sub}}$. These tasks reflect models' abilities on high-level concept abstraction, which aim to predict upper concepts multiple hops away in the concept hierarchy.

\subsection{Knowledge Concretization}
\label{sec:kcon}
Similar to KA and MKA tasks, this task contains KC (knowledge concretization) and MKC (multi-hop knowledge concretization) tasks.

\textbf{KC-Ins / KC-Sub:} KC-Ins and KC-Sub are head prediction tasks for ``\texttt{ins}'' and ``\texttt{sub}'' triples, which aim to predict entities for concepts or low-level concepts for high-level ones.

\textbf{MKC-Ins / MKC-Sub:} These tasks are head prediction tasks for multi-hop hierarchical triples. These tasks aim to predict entities for concepts or predict finer concepts for coarser concepts that are multi-hops away.

\subsection{Knowledge Completion}
The knowledge completion task contains the subtasks of entity graph completion (EGC) and concept graph completion (CGC) under two settings. In the ``Single'' setting, models can only use each single graph to do knowledge graph completion while both the two graphs and the cross-view links are provided in the ``Joint'' setting. % Combining the subtasks and settings, we have four subtasks and the details are described below.

\textbf{CGC-Single / EGC-Single}: These
% two
subtasks are conducted on each single graph $\mathcal{G}_C$ and $\mathcal{G}_E$.
% respectively. 
The test phase is conducted on logical triples of each graph $\mathcal{T}_E$ and $\mathcal{T}_{C(\texttt{logic})}$. The results can be compared with 
results from
CGC/EGC-Joint to {test} the effectiveness of jointly modeling the two graphs.

\textbf{CGC-Joint}: This subtask requires the model to do link prediction with the full information of the EC-KG $\mathcal{G}$. The model needs to abstract conceptual knowledge from the entity graph to do link prediction for logical concept graph triples $\mathcal{T}_{C(\texttt{logic})}$. The results of this subtask can also be used to verify models' abilities on knowledge abstraction.

\textbf{EGC-Joint}: Models are required to use the guidance from the concept graph to do link prediction for entity graph triples $\mathcal{T}_E$. For example, a person in the entity graph is more likely to lead some organizations if he is a politician. % Intuitively, the information from the concept graph can further help long-tail entities as models cannot handle them well because of the lack of corresponding triples. Link prediction results on long-tail entities can be further used to examine the concretization ability of existing models.

\section{Dataset Construction}
\begin{table*}[t!]
\begin{center}
\scalebox{0.7}{
\begin{tabular}{l r r r r r r r}
\toprule 
\multirow{2}{*}{\textbf{Dataset}} & \multicolumn{3}{c}{\textbf{Entity Graph}} & \multicolumn{3}{c}{\textbf{Concept Graph}} & \multirow{2}{*}{\textbf{\# Cross-links}}\\ \cline{2-4} \cline{5-7}
 & \textbf{\# Entities} & \textbf{\# Relations} & \textbf{\# Triples} &  \textbf{\# Concepts} & \textbf{\# Conceptual Rels} & \textbf{\# Triples} & \\ \midrule
YAGO26K-906 & 26,078 & 34 & 390,738 & 906 & 30 & 8,962 & 9,962 \\
DB111K-174 & 111,762 & 305 & 863,643 & 174 & 20 & 763 & 99,748 \\ \midrule
\name-S & 11,896 & 82 & 90,722 & 2,561 & 18 & 6,137 & 16,061\\
\name-M & 99,615 & 209 & 662,650 & 6,685 & 30 & 15,616 & 123,342\\
\name-L & 999,148 & 377 & 7,741,272 & 15,160 & 44 & 34,930 & 1,097,970 \\ % \hline 
\bottomrule
\end{tabular} }
\end{center}
\vspace{-0.7em}
\caption{\label{tab:dataset} Statistics of different datasets.}
\vspace{-1em}
\end{table*}

In this section, we first provide the details of our data collection process and annotation process. Then we give a detailed analysis of the statistical characteristics of the datasets.

\subsection{Dataset Collection}
The dataset construction process has four steps:
% and the details are described in the following.

\textbf{Step 1: Entity Filtering.}
We select entities in FB15K-237~\cite{toutanova2015representing} as our seed entities. We first find out corresponding seed entities in the Wikidata dump via the ``\texttt{FreebaseID}'' property of each item. Note some entities in Freebase may be labeled as concepts in Wikidata, so we filter out these concepts in our seed entity set.
Then we extract one-hop neighbors of the seed entities in the entity graph to form the entity pool, which contains more than 10 million entities.     
With the entity pool, we can sample an arbitrary size of one-hop neighbors to form the entity graph of our dataset. Our sampling strategy is to select entities with highest degrees and the final entity set consists of all seed entities and the sampled one-hop neighbors. To meet the requirements for different scales, we propose three sizes of datasets: (1) \name-S, the dataset only contains the seed entities; (2) \name-M, the expected total entity number is set to 100K; (3) \name-L, the entity number is set to 1M.
 
 \textbf{Step 2: Concept Finding.}
Next, we extract 
% corresponding
concepts based on selected entities. We use a breadth-first search algorithm to find the concepts. The algorithm starts from entities and search for concepts via ``\texttt{ins}'' triples and ``\texttt{sub}'' triples. Since the concept hierarchy follows the structure of a directed acyclic graph, our algorithm ends when all potential concepts are found.

\textbf{Step 3: Triple Extracting and Filtering.}
% With the selected entity set and concept set, 
This step  firstly extracts cross-view links and all triples in the entity/concept graph. Then we filter all triples by relation statistics and annotation. 
Relations (1) with less than 10 triples, (2) whose head or tail entity set's size is smaller than 10, and (3) which are annotated meaningless
% Relations with less than 10 triples, relations whose head or tail entity set size is less than 10, and relations which are annotated meaningless 
% are filtered out. 
are dropped. Similar to~\newcite{toutanova2015representing}, we further remove reverse relations to prevent valid/test leakage.

\textbf{Step 4: Concept Filtering.}
To get more precise concept graphs, we ask human annotators to find out meaningless concepts 
% in our datasets
and we further remove these concepts. These ``meaningless'' concepts include concepts with no labels or descriptions, concepts used for the self-construction of Wikidata (e.g. ``\textit{Wikimedia list article}''), etc. Details of the annotation step can be found in Appendix~\ref{sec:cpt-anno-appendix}. % todo

%   \begin{tikzpicture}
%     \draw (0,0) arc[radius=5pt,start angle= 90,end angle=270];
%     \draw (0,0) rectangle (40pt,-20pt);
%     \draw (40pt,0) arc[radius=5pt,start angle=90,end angle=-90];
%   \end{tikzpicture}
%   \begin{tikzpicture}
%     \node (1) [draw, rounded rectangle] {rounded rectangle};
%     \node (2) [below=of 1, draw, rounded rectangle, rounded rectangle west arc=0pt] {rounded rectangle};
%     \node (3) [below=of 2, draw, rounded rectangle, rounded rectangle east arc=0pt] {rounded rectangle};
%   \end{tikzpicture}
\subsection{Multi-hop Triple Annotation}
\label{sec:annotation}
% \begin{figure}
%     \centering
%     \includegraphics[width=0.9\linewidth]{acl-ijcnlp2021-templates/transitivity.pdf}
%     \caption{Examples of the concept hierarchy in YAGO26K-906 (top) and KACC (bottom).}
%     \label{fig:transitivity}
% \end{figure}
% \usetikzlibrary{shapes.misc, positioning}
% \node (1) [draw, rounded rectangle]
% \newcommand\mlnode[1]{\node (1) [draw, rounded rectangle]}
% \newcommand\mlnode[1]{\fbox{\begin{tabular}{@{}c@{}}#1\end{tabular}}}
% \newcommand\mlnode[1]{\fbox{\begin{tabular}{@{}c@{}}#1\end{tabular}}}
\newcommand\mlnode[1]{{\begin{tabular}{@{}c@{}}#1\end{tabular}}}
\begin{figure}
\centering
\scriptsize
\begin{tikzpicture}[node distance=1.8cm]
\node(writer)[roundrec]{\textit{writer}};
\node(person)[roundrec, right of=writer]{\textit{person}};
\node(organism)[roundrec, right of=person] {\textit{organism}};
\node(system)[roundrec, right of=organism] {\textit{system}};
\node(artist)[roundrec, above of=person, yshift=-1.2cm] {\textit{artist}};
\node(etc1)[draw=none, right of=artist, xshift=-0.4cm]{{...}};
\node(etc2)[draw=none, right of=system, xshift=-0.4cm]{{...}};
\node(name)[draw=none, above of=writer, yshift=-1.1cm, xshift=0.2cm]{YAGO26K-906:};
\draw[arrow](writer)--(artist);
\draw[arrow](writer)--(person);
\draw[arrow](person)--(organism);
\draw[arrow](organism)--(system);
\draw[arrow](system)--($(etc2.west)+(-0.1,0)$);
\draw[arrow](artist)--($(etc1.west)+(-0.1,0)$);
\draw[arrow]($(organism.east)+(-0.1,0.77)$)--($(system.west)+(0.2,0.77)$);
\node(legend)[draw=none, above of=system,xshift=0.3cm, yshift=-1.0cm]{\texttt{subclassOf}};
\end{tikzpicture}
\vspace{-0.5em}

\begin{tikzpicture}[node distance=1.8cm]
\node(scientist)[roundrec]{\textit{scientist}};
\node(researcher)[roundrec, right of=scientist]{\textit{researcher}};
\node(occupation)[roundrec, right of=researcher] {\textit{occupation}};
\node(humanactivity)[roundrec, minimum height=0.8cm,text depth = 0.09cm, right of=occupation, yshift=0.128cm] {\mlnode{\textit{human}\\\textit{activity}}};
\node(erudite)[roundrec, above of=researcher, yshift=-1.2cm] {\textit{erudite}};
\node(etc1)[draw=none, right of=erudite, xshift=-0.4cm]{{...}};
\node(etc2)[draw=none, right of=humanactivity, xshift=-0.4cm,yshift=-0.128cm]{{...}};
\node(name)[draw=none, above of=scientist, yshift=-1.1cm,xshift=-0.2cm]{KACC:};
\draw[arrow](scientist)--(erudite);
\draw[arrow](scientist)--(researcher);
\draw[arrow](researcher)--(occupation);
\draw[arrow](occupation)--($(humanactivity.west)+(0,-0.13)$);
\draw[arrow]($(humanactivity.east)+(0,-0.13)$)--($(etc2.west)+(-0.1,0)$);
\draw[arrow](erudite)--($(etc1.west)+(-0.1,0)$);
\end{tikzpicture}
% \caption{Examples of the concept hierarchy in YAGO26K-906 (top) and KACC (bottom). Arrows denote ``\texttt{subclassOf}'' relations.}
\vspace{-1.5em}
\caption{\footnotesize{Violation of  concept transitivity in two datasets.}}
% in YAGO26K-906 (top) and KACC (bottom). Arrows denote ``\texttt{subclassOf}'' relations.}
\vspace{-0.5cm}
\label{fig:transitivity}
\end{figure}
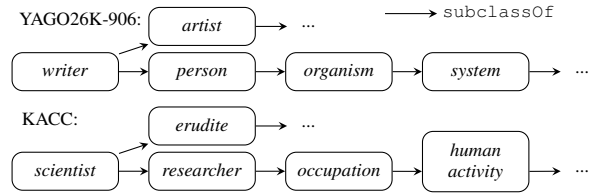

% \newcommand\mlnode[1]{{\begin{tabular}{@{}c@{}}#1\end{tabular}}}
% \begin{figure}
% \centering
% \scriptsize
% \begin{tikzcd}[row sep=-0.08cm, column sep = 0.44cm]
%  & \mlnode{\textit{artist}} & & & \\
%  \mlnode{\textit{writer}} \arrow[ru]\arrow[r] &  \mlnode{\textit{person}}\arrow[r] &  \mlnode{\textit{organism}} \arrow[r] &  \mlnode{\textit{system}} \arrow[r]& {...}
%  \end{tikzcd}
% \begin{tikzcd}[row sep=-0.2cm, column sep = 0.24cm]
%  & \mlnode{\textit{erudite}} & & & \\
%  \mlnode{\textit{scientist}}\arrow[ru]\arrow[r] &  \mlnode{\textit{researcher}}\arrow[r] &  \mlnode{\textit{occupation}} \arrow[r] & \mlnode{\textit{human}\\\textit{activity} }\arrow[r] & {...}
% \end{tikzcd}
% \vspace{-0.3cm}
% \caption{Examples of the concept hierarchy in YAGO26K-906 (top) and KACC (bottom). Arrows denote ``\texttt{subclassOf}'' relations.}
% \vspace{-0.5cm}
% \label{fig:transitivity}
% \end{figure}

To support MKA/MKC tasks, we extract multi-hop ``\texttt{ins}'' and ``\texttt{sub}'' triples from corresponding train sets according to rule~(\ref{equ:1}) and rule~(\ref{equ:2}). 
Ideally, hierarchical triples should preserve the multi-hop transitivity. However, when we dive into real-world datasets like YAGO26K-906 and ours, we find that the multi-hop transitivity does not always hold true. 
% For example.
As illustrated 
in Figure~\ref{fig:transitivity}, the transitivity is violated
% does not hold true 
when the transition link goes deep. 
% For example,
(\emph{scientist}, \texttt{sub}, \emph{occupation}) is meaningful while (\emph{scientist}, \texttt{sub}, \emph{human activity}) is not.
To make our multi-hop triples meaningful, we further 
% annotate the automaticly extracted multi-hop triples.  and 
ask human annotators to check the validity of these triples.  Details 
% of this annotation step
are in Appendix~\ref{sec:high-anno-appendix}.

\subsection{Dataset Analysis}
\label{sec:data_analysis}

In this subsection, we compare our datasets with existing datasets YAGO26K-906 and DB111K-174~\cite{hao2019universal} in terms of scale, domain coverage, and hierarchical relations. The statistics of these datasets are shown in Table~\ref{tab:dataset}.

\textbf{Scale.}
From Table~\ref{tab:dataset}, we can see that concept graphs in our three datasets have more balanced sizes compared to entity graphs.
From the comparison between DB111K-174 and \name, we can see that entity graphs of these two datasets have similar sizes, but {\name} has more concepts, conceptual relations, and triples in the concept graph.

Our datasets also have rich cross-view links. In Table~\ref{tab:dataset}, the average numbers of cross-links for each entity are less than 1.0 in YAGO26K-906 (0.38) and DB111K-174 (0.89), which means lots of entities in these datasets are not connected to concepts. In 
% our datasets
\name, the ratios are all 
% larger than 
above 1.09, indicating that one entity 
% in our dataset 
may belong to multiple concepts.

\begin{figure}[htbp!]
\vspace{-1em}
\centering
\subfigure{
\begin{minipage}[t]{0.44\linewidth}
\centering
\includegraphics[width=\linewidth]{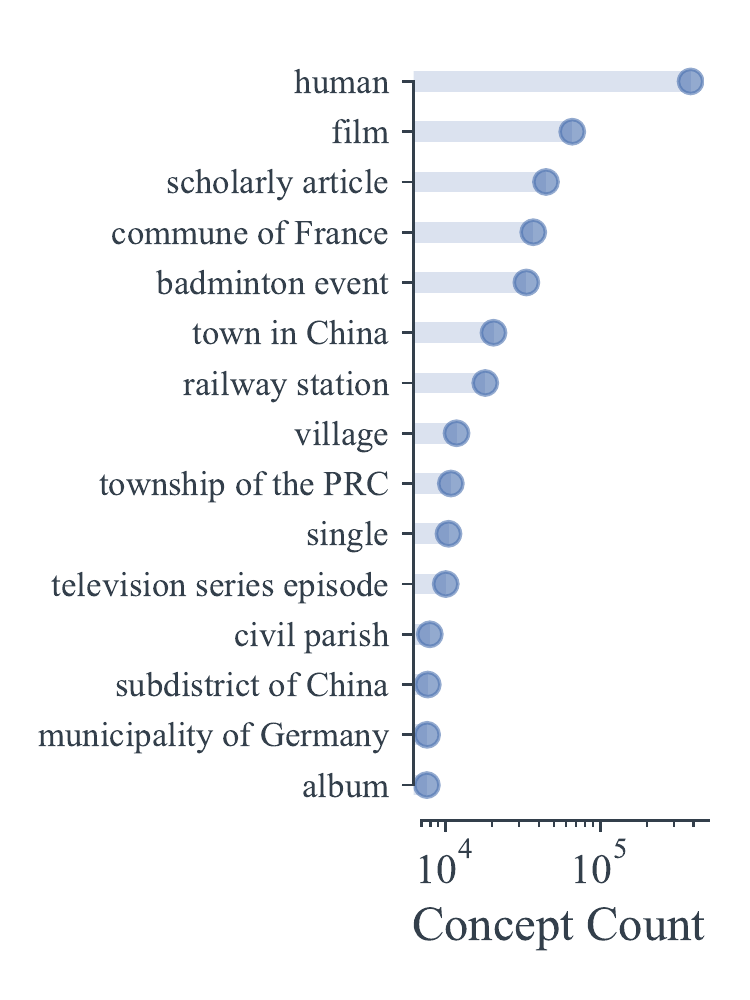}
%\caption{fig2}
\end{minipage}%
}%
% \subfigure{
% \begin{minipage}[t]{0.49\linewidth}
% \centering
% \includegraphics[width=\linewidth]{domain_KACC-Large.pdf}
% %\caption{fig2}
% \end{minipage}
% }%
\subfigure{
\begin{minipage}[t]{0.52\linewidth}
\centering
\includegraphics[width=\linewidth]{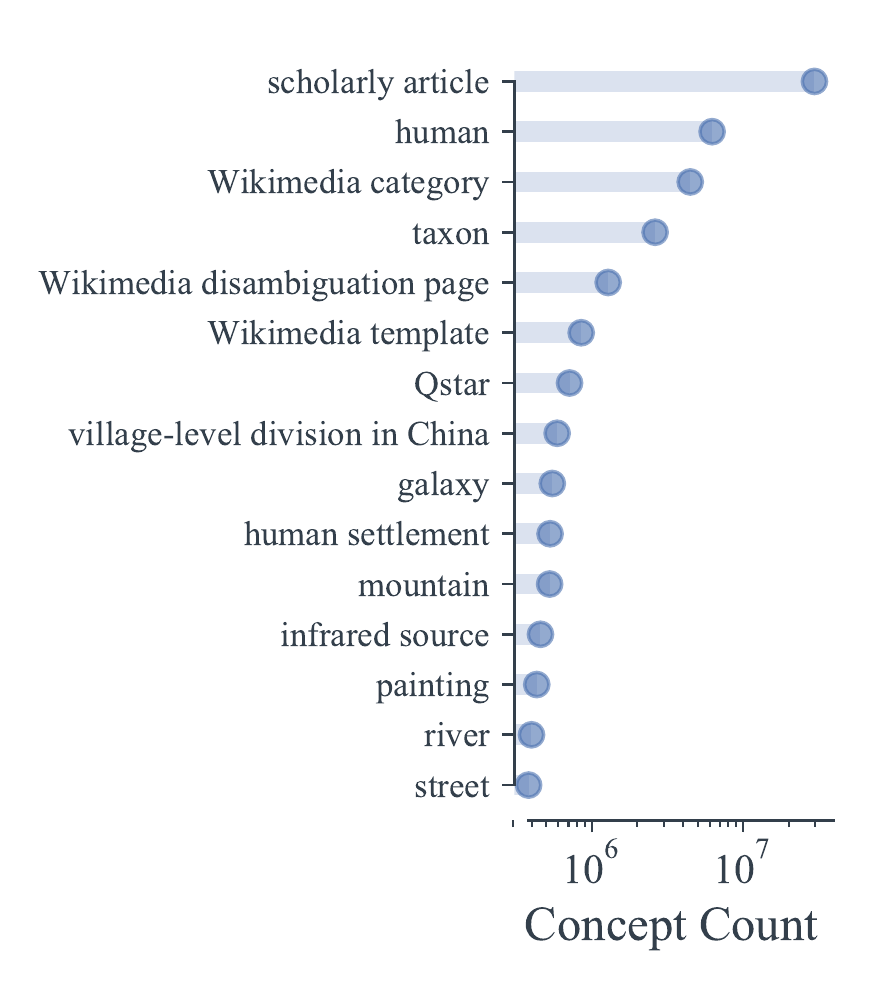}
%\caption{fig2}
\end{minipage}
}%
\centering
\vspace{-1em}
\caption{Top 15 most frequent bottom concepts of \name-L (left) and the original Wikidata dump (right).}
\label{fig:domain}
\end{figure}

\textbf{Domain Coverage.} In Figure~\ref{fig:domain}, we plot 15 most frequent bottom concepts, i.e., the concepts that directly connect to entities,  in \name-L and Wikidata dump to illustrate the domains of our datasets. Plots for \name-S and \name-M are in Appendix~\ref{sec:appendix-domain}. We find our datasets mainly focus on people, locations, sports, and films, similar to domains of our seed entities extracted from FB15K-237. 
% From KACC-S to KACC-L, more domains appear such as ``\emph{scholarly article}'', which is the most frequent concept in the original Wikidata dump (see the fourth subfigure). 
The comparisons between our datasets and Wikidata dump show that Wikidata dump contains more domains such as scholarly articles, galaxies, and entities related to Wikimedia. Our datasets only focus on partial domains of Wikidata dump, which ensures entities in our datasets are densely connected.

\textbf{Hierarchical Relations.} 
We present the characteristics of hierarchical relations in our datasets.

\begin{table}[]
\vspace{0.0em}
\newcommand{\tabincell}[2]{\begin{tabular}{@{}#1@{}}#2\end{tabular}}
    \centering
    \scalebox{0.7}{
    \begin{tabular}{l c c c} \toprule
    \textbf{Dataset} & \textbf{\tabincell{c}{\# Duplicate \\ Edges}} & \textbf{\# Self-loops}  & \textbf{\tabincell{c}{\# Undetected \\ Concepts}} \\ \midrule
     YAGO26K-906 & 188 & 44 & 132 (17.77\%)\\
     DB111K-174 & 27 & 13 & 4 (3.48\%)\\ \midrule
     KACC-S & 4 $\rightarrow$ 0 & 0 & 26 (1.02\%) $\rightarrow$ 0\\
     KACC-M & 13 $\rightarrow$ 0 &  0 & 33 (0.50\%) $\rightarrow$ 0\\
     KACC-L & 17 $\rightarrow$ 0 & 0 & 33 (0.22\%) $\rightarrow$ 0\\ \bottomrule
    \end{tabular}}
    \vspace{-0.4em}
    \caption{Quality check of existing datasets.``$\rightarrow$'' indicates the filtering process.}
    \label{tab:quality}
\vspace{-1em}
\end{table}

We first examine the data quality of \texttt{sub} triples in each dataset. We first detect duplicate edges and self-loops. As the global structure of \texttt{sub} relations is assumed to be a directed acyclic graph, we use the topological sort algorithm to find loops. We report numbers of concepts that are not detected by the algorithm in each dataset (these concepts are in loops or dangled in loops). The statistics are in Table~\ref{tab:quality}. We can see that our datasets are of high quality with fewer duplicate edges, no self-loops, and less proportion of concepts in loops. Finally, we remove duplicate edges and wrong-labeled triples in loops after a manual check.
% We manually check the loops and remove wrong-labeled triples in these loops.
\begin{figure}
\vspace{0em}
    \centering
    \includegraphics[width=0.7\linewidth]{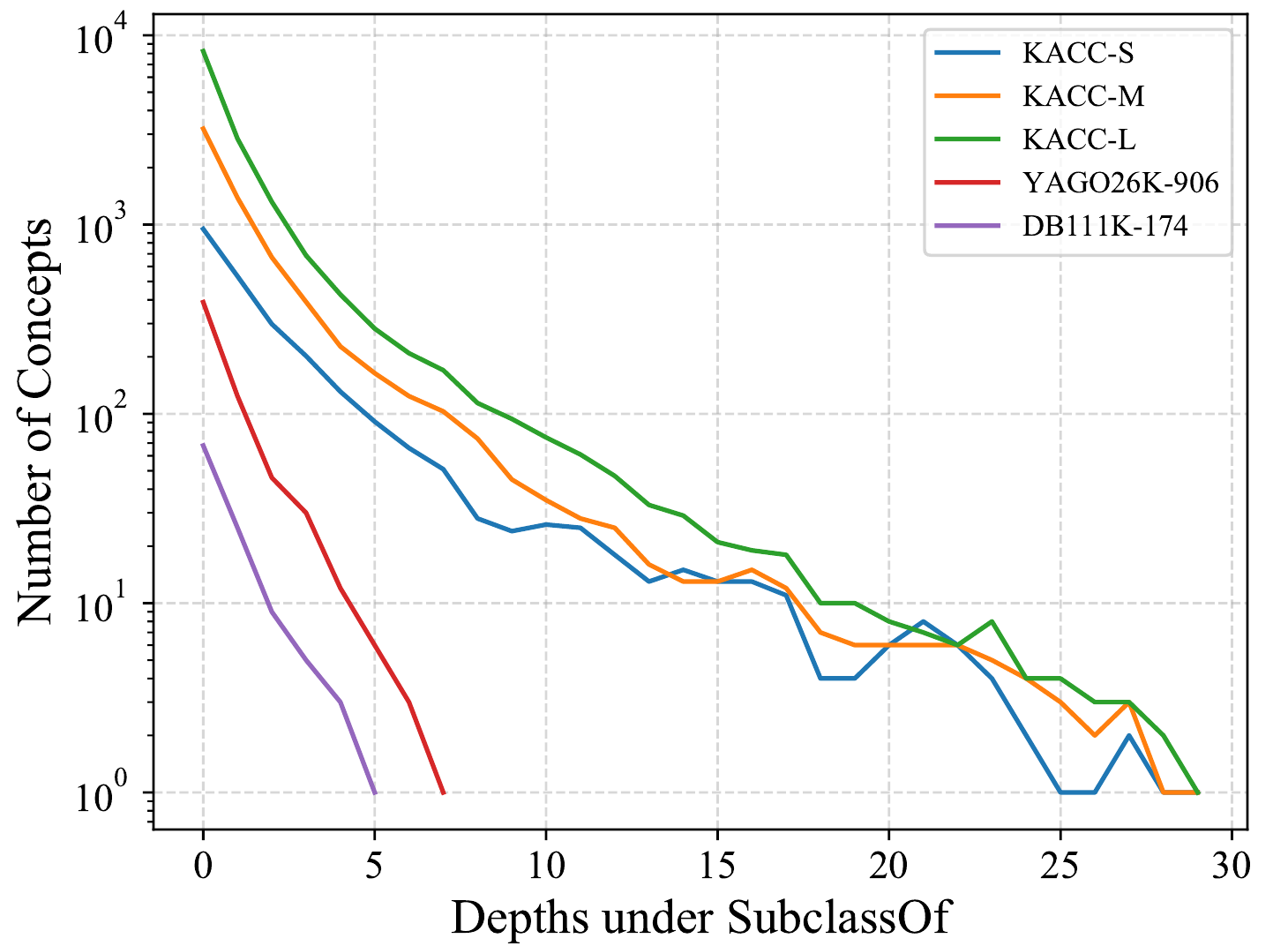}
    \vspace{-0.5em}
    \caption{Numbers of concepts with different depths.}
    \label{fig:depth}
\vspace{-1.5em}
\end{figure}

Then we examine the depths of the concepts in each dataset. We start from bottom concepts and traverse all concepts via topological sort. We plot numbers of concepts with different depths in Figure~\ref{fig:depth}. From the figure, we can see that our datasets have deeper hierarchical structures than others, which are more informative and useful for models to learn more fine-grained representations.

Finally, we show the characteristic of the ``\texttt{ins}'' relation in our datasets. Unlike existing datasets where ``\texttt{ins}'' only connects entities and concepts, concepts in Wikidata also have ``\texttt{ins}'' connections, which are denoted by $\mathcal{T}_{C(\texttt{ins})} $. We find these triples are also meaningful as they reflect different level semantics. For example in a triple (\emph{planet}, \texttt{ins}, \emph{astronomical object type}), ``\emph{planet}'' is a concept while it can also be regarded as an instance when mentioned in a higher level. The finding is also compatible with human cognition. So we remain these triples in our datasets, and we test them together with other ``\texttt{ins}'' triples. Therefore, we modify the corresponding definitions in Section~\ref{sec:formalization} into $\mathcal{T}_\texttt{ins} = \mathcal{T}_S\cup\mathcal{T}_{C(\texttt{ins})}$ and $\mathcal{T}_{C(\texttt{logic})}= \mathcal{T}_{C} \backslash (\mathcal{T}_{{\texttt{sub}}}\cup \mathcal{T}_{C(\texttt{ins})})$.

\textbf{Other Characteristics.}
Our datasets also have some new properties. In existing datasets, relations in the entity graph and conceptual relations in the concept graph are disjoint. However, in our datasets, some relations appear in both the entity graph and the concept graph. For example, the ``\texttt{partOf}'' relation appears in  {(\emph{Chile}, \texttt{partOf}, \emph{South America})} in the entity graph and {(\emph{hospital}, \texttt{partOf}, \emph{health system})} in the concept graph.
% We think this reflects the true data distribution in real KGs and keep these triples. 
Our experiments treat them as different relations while models can also treat these relations as the same, which depends on the hypotheses of the designers.

\section{Experimental Settings}
\begin{table}[t!]
\begin{center}
\scalebox{0.70}{
\begin{tabular}{c c c c }
\toprule 
\textbf{Task} & \textbf{Train} & \textbf{Valid} & \textbf{Test} \\
\midrule
 KA-Ins / KC-Ins & $\mathcal{T}_{EC}^{\text{Train}}$ & $\mathcal{T}_{\texttt{ins}}^{\text{Valid}}$ & $\mathcal{T}_{\texttt{ins}}^{\text{Test}}$\\ 
\rule{0pt}{12pt} KA-Sub / KC-Sub & $\mathcal{T}_{EC}^{\text{Train}}$ & $\mathcal{T}_{\texttt{sub}}^{\text{Valid}}$ & $\mathcal{T}_{\texttt{sub}}^{\text{Test}}$ \\
\rule{0pt}{12pt} MKA-Ins / MKC-Ins & $\mathcal{T}_{EC}^{\text{Train}}$ & $\mathcal{T}_{\texttt{M-Ins}}^{\text{Valid}}$ & $\mathcal{T}_{\texttt{M-Ins}}^{\text{Test}}$ \\
\rule{0pt}{12pt} MKA-Sub / MKC-Sub & $\mathcal{T}_{EC}^{\text{Train}}$ & $\mathcal{T}_{\texttt{M-Sub}}^{\text{Valid}}$ & $\mathcal{T}_{\texttt{M-Sub
}}^{\text{Test}}$  \\
\midrule
EGC-Joint & $\mathcal{T}_{EC}^{\text{Train}}$ & $\mathcal{T}_E^{\text{Valid}}$ &  $\mathcal{T}_E^{\text{Test}}$ \\
\rule{0pt}{12pt}CGC-Joint & $\mathcal{T}_{EC}^{\text{Train}}$ & $\mathcal{T}_{C(\texttt{logic})}^{\text{Valid}}$ & $\mathcal{T}_{C(\texttt{logic})}^{\text{Test}}$ \\
\rule{0pt}{12pt}EGC-Single & $\mathcal{T}_E^{\text{Train}}$ & $\mathcal{T}_E^{\text{Valid}}$ & $\mathcal{T}_E^{\text{Test}}$ \\
\rule{0pt}{12pt}CGC-Single & $\mathcal{T}_C^{\text{Train}}$ & $\mathcal{T}_{C(\texttt{logic})}^{\text{Valid}}$ & $\mathcal{T}_{C(\texttt{logic})}^{\text{Test}}$ \\
\bottomrule
\end{tabular}}
\end{center}
\vspace{-0.5em}
\caption{\label{tab:task-definition} Settings of datasets for different tasks.}
\vspace{-1.5em}
\end{table}

\begin{table*}[t!]
\begin{center}
\scalebox{0.7}{
\begin{tabular}{c |c c c c c c| c c c c c c }
\Bhline 
\multirow{2}{*}{\textbf{Method}} & \multicolumn{3}{c}{\textbf{KA-Ins}} & \multicolumn{3}{c|}{\textbf{MKA-Ins}} & \multicolumn{3}{c}{\textbf{KA-Sub}} & \multicolumn{3}{c}{\textbf{MKA-Sub}} \\ \cline{2-13} 
& \textbf{MRR} & \textbf{Hits@1} & \textbf{Hits@10} & \textbf{MRR} & \textbf{Hits@1} & \textbf{Hits@10}& \textbf{MRR} & \textbf{Hits@1} & \textbf{Hits@10}& \textbf{MRR} & \textbf{Hits@1} & \textbf{Hits@10} \\  \hline 
TransE & 0.658 & 0.560 & 0.832 &0.112  &0.047  & 0.242 & 0.093 & 0.000 & 0.288 & 0.098 &0.035&0.225\\
DistMult & 0.712 & 0.636 & 0.847 &0.131  & 0.086 & 0.222 & 0.135 &0.062 & 0.277 & 0.122&0.040&0.284\\
ComplEx & 0.737 & 0.663  & 0.863 & 0.121 & 0.078 &0.214  & \textbf{0.226} & \textbf{0.151} & 0.373 & 0.135 & 0.066&0.275\\
TuckER & 0.759 & 0.681 & 0.885 &0.115  & 0.077 &0.179 & 0.191  &0.107 & 0.369 &0.147 & 0.068 & 0.313\\ \hline
JOIE & 0.706 & 0.611 & 0.873 & 0.195 & 0.115 & 0.370 & 0.099 & 0.004 & 0.289 & 0.113 & 0.010 & 0.351\\
AttH & \textbf{0.778} & \textbf{0.693} & \textbf{0.918} & \textbf{0.218} & \textbf{0.116} & \textbf{0.436} & 0.203 & 0.089 & \textbf{0.458} & \textbf{0.188} & \textbf{{0.081}} & \textbf{0.420} \\
\Bhline
\end{tabular} }
\end{center}
\vspace{-1em}
\caption{\label{tab:ka} Results on knowledge abstraction. Best scores are in \textbf{bold}.}
\vspace{-0.5em}
\end{table*}

\begin{table*}[t!]
\begin{center}
\scalebox{0.7}{
\begin{tabular}{c |c c c c c c| c c c c c c}
\Bhline 
\multirow{2}{*}{\textbf{Method}} & \multicolumn{3}{c}{\textbf{KC-Ins}} & \multicolumn{3}{c|}{\textbf{MKC-Ins}} & \multicolumn{3}{c}{\textbf{KC-Sub}} & \multicolumn{3}{c}{\textbf{MKC-Sub}} \\ \cline{2-13} 
& \textbf{MRR} & \textbf{Hits@1} & \textbf{Hits@10} & \textbf{MRR} & \textbf{Hits@1} & \textbf{Hits@10}& \textbf{MRR} & \textbf{Hits@1} & \textbf{Hits@10}& \textbf{MRR} & \textbf{Hits@1} & \textbf{Hits@10} \\  \hline
TransE & 0.123 & 0.083 & 0.208 & 0.060&0.033  &0.110  & 0.049& 0.000  &0.145    & 0.039 &0.000&0.113\\
DistMult & 0.175 & 0.114 & 0.282 &0.107  & 0.068 & 0.183 & 0.061 &0.021  &0.136  &0.041  &0.004&0.128\\
ComplEx & {0.208} & {0.142} & \textbf{0.321} & 0.098 & {0.070} & 0.153 & \textbf{0.103} & \textbf{0.061} & 0.179 & 0.044 & {0.012} & 0.107\\
TuckER & 0.169 & 0.110 & 0.280 &  0.074 & 0.047 & 0.120 & 0.087  &0.040  & 0.160 &\textbf{0.051} & \textbf{0.021}&0.106\\ \hline
JOIE & \textbf{0.241} & \textbf{0.200} & 0.320 & \textbf{0.141} & \textbf{0.093} & \textbf{0.240} & 0.048 & 0.010 & 0.120 & 0.032 & 0.001 & 0.092 \\
AttH & 0.172 & 0.120 & 0.279 & {0.112} & 0.061 & {0.213} & 0.081 & 0.021 & \textbf{0.204} & {\textbf{0.051}} & 0.009 & \textbf{0.135}\\
\Bhline
\end{tabular} }
\end{center}
\caption{\label{tab:kcon} Results on knowledge concretization. Best scores are in \textbf{bold}.}
\vspace{-1em}
\end{table*}

\subsection{Dataset Partition}
Considering the tradeoff between scale and training efficiency, we use the medium-sized dataset KACC-M to conduct the experiments. 
% We first split cross-view links $\mathcal{T}_S$, triples in the entity graph $\mathcal{T}_E$ and triples in the concept graph $\mathcal{T}_C$ by the proportion 8:1:1.
To generate each task's train/valid/test data, we firstly split each triple set $\mathcal{T}_S, \mathcal{T}_E$ and $\mathcal{T}_C$ by the proportion 8:1:1. 
% Then we use these splits according to Table~\ref{tab:task-definition}.
% Because of the difference between each task, we enumerate the train/valid/test set used by each task  /\in
% Our split algorithm ensures that are no unseen entities or relations in the valid/test set. 
% After the partition, we can get corresponding sets $\mathcal{T}_S^{\text{Train}} / \mathcal{T}_S^{\text{Valid}} / \mathcal{T}_S^{\text{Test}}$, $\mathcal{T}_E^{\text{Train}} / \mathcal{T}_E^{\text{Valid}} / \mathcal{T}_E^{\text{Test}}$, and $\mathcal{T}_C^{\text{Train}} / \mathcal{T}_C^{\text{Valid}} / \mathcal{T}_C^{\text{Test}}$. 
%For HC and TC sets, the data is sparse and we only use them as test sets. 
 % The statistics of these datasets are shown in Table~\ref{tab:data-split}.
% Previous methods use different train sets for different tasks based on a same EC-KG.
% Since our goal is to simultaneously embed entities and concepts, as well as model the logical triples and the concept hierarchy at the same time, we use all train splits $\mathcal{T}_{EC}^{\text{Train}} = \mathcal{T}_E^{\text{Train}} \cup \mathcal{T}_C^{\text{Train}} \cup \mathcal{T}_S^{\text{Train}}$ to train our model. 
To make it easy for model training and hyper-parameter selection, we provide a unified train set $\mathcal{T}_{EC}^{\text{Train}}$ for all tasks defined on the EC-KG, that is $\mathcal{T}_{EC}^{\text{Train}} = \mathcal{T}_E^{\text{Train}} \cup \mathcal{T}_C^{\text{Train}} \cup \mathcal{T}_S^{\text{Train}}$.
%The train set $\mathcal{T}_{EC}^{\text{Train}}$ contains the train sets from the entity graph, concept graph and cross-view links, that is, $\mathcal{T}_{EC}^{\text{Train}} = \mathcal{T}_E^{\text{Train}} \cup \mathcal{T}_C^{\text{Train}} \cup \mathcal{T}_S^{\text{Train}}$. 
For the valid and test sets, different tasks have their own valid/test sets for model selection and performance reports.

Train sets are different for EGC-Single and CGC-Single.
As they focus on a single graph, we use $\mathcal{T}_E^{\text{Train}} / \mathcal{T}_C^{\text{Train}}$ as train sets respectively. 
%  We use $\mathcal{T}_E^{\text{Train}} / \mathcal{T}_C^{\text{Train}}$ as train sets, respectively since they focus on a single graph
The settings of datasets for different tasks are in Table~\ref{tab:task-definition}. The statistics are in the Appendix~\ref{sec:data-appendix}.
% For EGC-Single and CGC-Single tasks, their train sets are different. As these tasks focus on a single graph, we use $\mathcal{T}_E^{\text{Train}} / \mathcal{T}_C^{\text{Train}}$ as train sets, $\mathcal{T}_E^{\text{Valid}} / \mathcal{T}_C^{\text{Valid}}$ as valid sets and $\mathcal{T}_E^{\text{Test}} / \mathcal{T}_C^{\text{Test}}$ as test sets for these two tasks respectively. The settings of datasets for different tasks are shown in Table~\ref{tab:task-definition}. 

\subsection{Baselines}
To test how existing methods behave in our benchmark, we choose several representative models for single-view KG embedding, as well as JOIE~\cite{hao2019universal} and AttH~\cite{chami2020low},
% Our baselines contain several knowledge embedding methods as well as JOIE,
which are specially designed for modeling the EC-KG.

\textbf{Single-view KE Methods.} We use TransE~\cite{bordes2013translating}, DistMult~\cite{yang2015embedding}, ComplEx~\cite{trouillon2016complex}, and TuckER~\cite{balavzevic2019tucker} as our baselines. These baselines treat the EC-KG as a large single-view KG by regarding concepts as entities, conceptual relations and hierarchical relations as ordinary relations defined on a single-view KG. 

\textbf{JOIE.} JOIE~\cite{hao2019universal} uses traditional KE methods like TransE and DistMult as the backend model to learn logical relations in entity/concept graphs. It further defines specific transformations and loss functions for hierarchical triples. These mechanisms could improve the performance of corresponding backend models. % It further defines two transformations ``Cross-view Grouping (CG)'' and ``Cross-view Transformation (CT)'' for cross-view links. The loss for CG directly minimizes the distances between entities and concepts while in the CT setting, entities are first transformed to the embedding space of concepts, then the distances between entities and concepts are computed. 

\textbf{AttH.} AttH~\cite{chami2020low} utilizes the hyperbolic geometry to embed  tree-like structures, which is suitable for modeling the concept hierarchy.  It also proposes methods to embed logical relations in the hyperbolic space. % AttH achieves SOTA MRRs on WN18RR and YAGO3-10. 
% Note that ~\citet{chami2020low} proposes three variants in their work with similar performance. We choose AttH because it achieves better performance in FB15k-237 which is  similar to \name \  in terms of domain.
% It also propose several mechanism to model the logic relations uses hyperbolic embedding to capture concept hierarichy.
% \textbf{TransC.} TransC~\cite{lv2018differentiating} tries to differentiates concepts and entities. It uses TransE to model the entity graph and it further encodes each concept as a sphere with a vector $\textbf{c}$ as the position of its center and a scalar $r$ as its radius. The idea is that entities belonged to a concept should be located inside the sphere of the concept in the embedding space. TransC is designed for the datasets where the concept hierarchy has a tree structure without meta-relations. So that TransC cannot utilize the information from meta-relations and it cannot deal with the CGC-Joint task.

\subsection{Evaluation Metrics}
We test the tasks in the form of link prediction. We use two evaluation metrics in these tasks:

\textbf{Mean Reciprocal Rank (MRR).} The metric computes the mean reciprocal rank of the correct instances. If the ranks of correct instances are ${k_i}$, then the metric computes the average of $\frac{1}{k_i}$.

\textbf{Hits@N.} This metric computes the proportion of the ranks that are no larger than N.

A good model could achieve higher scores on these metrics. We use the ``Filtered'' setting for all the evaluations, which filters out other true answers from the prediction results to get the final rank for each test case.

\subsection{Hyperparameter Settings}
According to  ~\citet{ruffinelli2020you}, performances of KGE methods are sensitive to hyperparameters. Following them, we run 30 quasi-random trails for all models from predefined hyperparameter spaces. We list the hyperparameter spaces we use in Appendix ~\ref{sec:param-space-appendix}. We run all trails for 100 epochs.

For all single-view KE methods, we use the implementations from LibKGE~\cite{libkge}, which utilizes the Ax framework to perform quasi-random hyperparameter search.  % For other parameters, we follow the best parameters provided by OpenKE on the FB15K-237 dataset. 

For AttH, we use the implementation from the authors\footnote{{{https://github.com/HazyResearch/KGEmb}}}. % For a fair comparison, we use the same hyperparameter space with KE methods.
% Since TransE and AttH only support negative sampling as the training scheme~\cite{ruffinelli2020you}, the hyperparameter space for them do not include the ``1vsAll'' scheme.
For JOIE, we use the implementation from the authors\footnote{{{https://github.com/JunhengH/joie-kdd19}}}. We use TransE as the backend and adopt the suggested hyperparameter space from the paper.

% For TransC, we use the implementation from the authors' Github\footnote{https://github.com/davidlvxin/TransC}. We follow the best parameters from the paper and select the embedding dimension from \{100, 200, 300\} based on the valid MRR. The model is also trained for 100 epochs.

\begin{table*}[!htbp]
\begin{center}
\definecolor{light-gray}{gray}{0.85}
\newcolumntype{g}{>{\columncolor{light-gray}}c}
% \newcolumntype{g}{>{\columncolor{myLightCyan}}c}
\scalebox{0.7}{
\begin{tabular}{c |g c g c g c| g c g c g c}
\Bhline
\multirow{3}{*}{\textbf{Method}} & \multicolumn{6}{c| }{\textbf{EGC}} & \multicolumn{6}{c }{\textbf{CGC}}  \\ \cline{2-13} 
& \multicolumn{2}{c}{\textbf{MRR}} &  \multicolumn{2}{c}{\textbf{Hits@1}} & \multicolumn{2}{c|}{\textbf{Hits@10}} & \multicolumn{2}{c}{\textbf{MRR}} &  \multicolumn{2}{c}{\textbf{Hits@1}} & \multicolumn{2}{c}{\textbf{Hits@10}} \\  \cline{2-13}
& \textbf{Joint} & \textbf{Single} & \textbf{Joint} & \textbf{Single} & \textbf{Joint} & \textbf{Single} & \textbf{Joint} & \textbf{Single} & \textbf{Joint} & \textbf{Single} & \textbf{Joint} & \textbf{Single} \\  \hline
TransE & 	\underline{0.305} & 0.299 & \underline{0.182} & 0.175 &\underline{0.510} & 0.504 & 0.242 & \underline{0.261} & 0.003 & \underline{0.095} & \underline{0.659} & 0.603 \\
DistMult & \underline{0.444} & 0.440 & \underline{0.379} & 0.376 & \underline{0.566} & 0.560 & \underline{0.495} & 0.481 & 0.419 & \underline{0.436} & \underline{0.631} & 0.568 \\
ComplEx &\underline{0.458} & 0.453 & \underline{0.397} & 0.393 & \underline{0.572} & 0.567 & \underline{\textbf{0.537}} & 0.501 &\underline{\textbf{0.472}} & 0.458 & \underline{\textbf{0.684}} & 0.581\\ 
TuckER  &\underline{\textbf{0.481}} & \textbf{0.473} & \underline{\textbf{0.415}} & \textbf{0.408} & \underline{\textbf{0.604}} & \textbf{0.595} & \underline{0.536} & \textbf{0.525} & 0.468 & \underline{\textbf{0.473}} & \underline{0.668} & \textbf{0.615} \\ \hline
JOIE & 0.171 & - & 0.094 & - & 0.308 & - & 0.218 & - & 0.018 & - & 0.622 & -
 \\
AttH & 0.348 & \underline{0.352} & 0.235 & \underline{0.241} & \underline{0.551} & 0.545 & \underline{0.268} & 0.244 & 0.100 & \underline{0.154} & \underline{0.631} & 0.418 \\ \Bhline
\end{tabular} }
\end{center}
\vspace{-1em}
\caption{\label{tab:com} Results on knowledge completion. Best scores among different models in the same task are in \textbf{bold}. Best scores for a model between Joint and Single settings are \underline{underlined}.}
\vspace{-1em}
\end{table*}

\section{Experimental Results}
In this section, we provide the experimental results
% of different baselines 
and further propose several future directions.
\subsection{Knowledge Abstraction}
The results of knowledge abstraction are shown in Table~\ref{tab:ka}. From the results, we can see that AttH  has a large margin beyond other methods on KA-Ins and also performs well on KA-Sub, which demonstrates the effectiveness of hyperbolic embeddings. JOIE outperforms its backend model TransE.

% Comparing results between KA-Ins / KA-Sub and MKA-Ins / MKA-Sub, we find that almost all methods have large performance drops on multi-hop triples. The drops on KA-Ins and MKA-Ins are much larger than KA-Sub and MKA-Sub, which shows that multi-hop ``\texttt{instanceOf}'' triples are much harder for existing methods. Among these methods, performances of AttH drops least, which shows that hyperbolic methods are more suitable to model multi-hop transitivity.

Comparing results between KA-Ins and MKA-Ins, all the models have performance degradation larger than 0.51 on MRR. We conclude that the composition rule in Equation~(\ref{equ:1}) is hard to learn naturally. Among all the models, AttH performs the best on both tasks and has the least degradation from KA to MKA, 
% which shows 
showing
that hyperbolic space has advantages over Euclidean space in 
% terms of 
knowledge abstraction. However, the degradation is still drastic, showing the difficulty of the MKA task.

Comparing results between KA-Sub and MKA-Sub, most methods also have performance degradation while TransE-based models (TransE and JOIE) have better performances on MKA-Sub, which is interesting for further investigation. AttH performs best on MKA-Sub, which further confirms the advantage of hyperbolic methods. %  However, the performance in KA-Sub is much lower than results on logical relations in Table~\ref{tab:com}, demonstrating the challenges of modeling ``\texttt{subclassOf}'' relation.

\subsection{Knowledge Concretization}

The results of knowledge concretization are in Table~\ref{tab:kcon}. ComplEx and JOIE performs well on KC-Ins and KC-Sub tasks. Similar to tasks in knowledge abstraction, MKC-Ins and MKC-Sub are also harder for existing models. 
%AttH has better results on multi-hop triples, which is consistent with the analysis in knowledge abstraction. 
The results of knowledge concretization tasks are lower than corresponding knowledge abstraction tasks, which shows that knowledge concretization is much harder than knowledge abstraction.

\subsection{Knowledge Completion}
The results of knowledge completion are shown in Table~\ref{tab:com}. From the table, TuckER performs well on entity-level logical relations while ComplEx is good at dealing with concept-level logical relations. JOIE does not perform well on logical relations.

From the comparisons between ``Joint'' and ``Single'' settings, we find that results on EGC-Joint are usually higher than results on EGC-Single, which shows that incorporating the concept graph and cross-view links helps the understanding of the entity graph. However, the pattern is not obvious on CGC-Joint and CGC-Single, which may due to that entity triples are far more than concept triples, so models tend to focus more on entity triples.

\subsection{Overall Results}

Finally, we compute an overall KACC score for each method to show their overall performances. Similar to GLUE~\cite{wang2018glue}, we average Hits@10 scores of each method on all tasks (except CGC-Single and EGC-Single) to get final scores. We also compute the average scores for knowledge abstraction (KA), knowledge concretization (KCon), and knowledge completion (KCom). In Table~\ref{tab:kacc-score} we can see that 
AttH has the best overall score and achieves the highest scores on two tasks. ComplEx also performs well. It is a balanced model since it gets the second place on all tasks. TuckER performs best on knowledge completion. In the future, we plan to test more methods and investigate their abilities.

\begin{table}[t!]
\vspace{0.4em}
\begin{center}
\scalebox{0.7}{
\begin{tabular}{c c c c c}
\toprule 
\textbf{Method} & \textbf{KACC} & \textbf{KA} & \textbf{KCon} & \textbf{KCom}\\  \midrule 
TransE &  0.374 & 0.396 & 0.143  & 0.585  \\
DistMult & 0.396 & 0.407 & 0.181 & 0.599 \\
ComplEx & \underline{0.414} & \underline{0.429} & \underline{0.184} & \underline{0.628} \\
TuckER & 0.410 & 0.423 & 0.165  & \textbf{0.636}\\
JOIE & 0.353 & 0.444 & 0.177 & 0.439 \\
AttH & \textbf{0.452} & \textbf{0.558} & \textbf{0.208} & 0.591\\
\bottomrule
\end{tabular} }
\end{center}
\vspace{-1em}
\caption{\label{tab:kacc-score} The overall scores. Best scores are in \textbf{bold} and second high scores are \underline{underlined}. }
\vspace{-1em}
\end{table}

\subsection{Analyses and Future Directions}
From the results above, we analyze several problems that existing models cannot handle well and propose several promising future directions.

\textbf{Multi-hop triple modeling.} The prediction scores of multi-hop triples are lower than those of one-hop triples, showing the challenge of  multi-hop triple modeling. Besides, how to balance the model to learn from logical and hierarchical relations is also an exciting direction.

% The underlying reason is that ``\texttt{subclassOf}" triples form a hierarchy and the triples preserve the transitivity. For example, $(C, \texttt{subclassOf}, A)$ will be true if we have  $(C, \texttt{subclassOf}, B)$ and  $(B, \texttt{subclassOf}, A)$. But existing models cannot deal with the transitivity. Consequently, techniques proposed for hierarchy modeling are encouraged, such as modeling concepts as spheres~\cite{lv2018differentiating} or using hyperbolic embeddings~\cite{chami2020low}.
     
\textbf{Conceptual knowledge completion.} Not all models successfully extract conceptual knowledge effectively as their scores of CGC-Joint are lower than those of CGC-Single. The main reason is that KE methods tend to focus more on entity triples 
due to the losses.
% because of the design of the losses.
They lack the ability to abstract factual knowledge to enrich conceptual knowledge. 
% Some improvements may be triple reweighting or finding important factual triples.

\iffalse
\begin{figure}
    \centering
    \includegraphics[width=0.8\linewidth]{case.pdf}
    \caption{A case of entities and concepts in the embedding space. In this case, entities of concept A and concept B can find correct concepts based on distances, but concept A cannot find its correct entities.}
    \label{fig:case}
\end{figure}
\fi

\textbf{Knowledge concretization.} The results of concretization tasks are much lower than those of abstraction tasks. It demonstrates that existing models can find proper concepts for entities but cannot find correct entities for concepts. 
% A simple case of this phenomena is shown in Figure~\ref{fig:case}. 
Some solutions may be 
% selecting hard negative samples or 
using contrastive learning to ``push" negative entities away from the concepts. 

Besides the analyses, there are also several promising future directions of our benchmark.

\textbf{Contextualized knowledge embedding.} Recently, contextualized knowledge embeddings~\cite{wang2019coke} are proposed to capture different semantics of entities and relations in different contexts. These methods only conduct on the entity graph, while incorporating concepts provides more contextual information for entities. For example, an entity of a \emph{painter} is more likely to $\texttt{paint}$ than a \emph{politician}. 
It is a promising direction to model concepts and entities jointly by contextualized embeddings. 

\textbf{Joint modeling EC-KG with text.} The EC-KG is a symbolic form of knowledge, and it is interesting to combine it with text. Future directions may include incorporating more commonsense knowledge into the concept graph from language or using the EC-KG to help understand the natural language from the concept level.

\iffalse
\textbf{New task settings.} Researchers are also encouraged to propose new task settings based on the EC-KG. For example, EI only predicts direct descendants for bottom concepts, a more challenging entity inference task can be defined for upper-level concepts, which have no direct links to entities. 
\fi

\section{Conclusion}
In this paper, we focus on the problems of knowledge abstraction, concretization, and completion. We propose a benchmark to test the abilities of models on KACC. To conduct the evaluation, we construct large-scale datasets with desired properties, and experiments show that tasks in KACC are challenging.
For future work, we plan to test more models and design advanced models to address tasks in  KACC.
% KACC is an open benchmark, and we encourage researchers to propose new meaningful tasks based on the nature of the EC-KG.
%Inspired by the experiments, we propose several interesting future research directions: (1) Better modeling of the ``\texttt{isA}'' relation; (2) Capturing the contextualized property provided by the abundant cross-view links in KACC; (3) Finding useful prototypes for concepts; (4) Incorporating more commonsense knowledge into the concept graph from language or other KGs and using the EC-KG to help the understanding of language.

\section{Acknowledgements}
This work is funded by the Natural Science Foundation of China (NSFC) and the German Research Foundation (DFG) in Project Crossmodal Learning, NSFC 62061136001 / DFG TRR-169. This work is also supported by Tencent Marketing Solution Rhino-Bird Focused Research Program.

\section*{Ethical Considerations}
Here we list ethical considerations of our paper:

\textbf{Intellectual property.} All of our datasets are collected from Wikidata and Wikidata offers the data for free with no requirement to attribute under \href{https://www.wikidata.org/wiki/Wikidata:Text_of_the_Creative_Commons_Public_Domain_Dedication}{Creative Commons CC0 License}. 

\textbf{Privacy.} Our datasets are collected from an online resource automatically and the collection process does not involve with participants' privacy rights.

\textbf{Compensation.} For the two annotation processes, the salary for annotating each sample is computed according to the average annotation time and local wage standard. And we ensure that all annotators are well paid.

\textbf{Potiential problems.} Though we have manually checked the quality of our datasets and removed meaningless and wrong data, there still may exist false triples. These may lead to wrong predictions in knowledge abstraction, concretization and completion tasks. However, noises are common in human contributed resources such as existing datasets and ours, so the potiental risks are low.

\bibliographystyle{acl_natbib}
\bibliography{acl2021.bib}

\input{appendix.tex}

\end{document}

%% file: appendix.tex
\newpage
\phantom{.}
\newpage
\appendix
\section{Appendices}
\subsection{Annotations for Meaningless Concepts}
\label{sec:cpt-anno-appendix}

In this section, we first present our annotation guidelines for annotators, and then we provide the annotation results. 

\paragraph{Task Guidelines.} This task aims to find out meaningless ``concepts''. For a given instance, you need to check whether it is a ``concept''. Here are some definitions in this task:
\begin{itemize}
    \item \textbf{Concept.} A word for a group or a class of things, such as ``artist'', ``writer'', etc. Humans obtain concepts by abstracting commonalities from things.
    \item \textbf{Entity.} A specific person or thing, such as ``Barack Obama'', ``Mona Lisa'', etc.  
\end{itemize}
We provide you the Wikidata ID, name, and description of an instance. For more details, you can go to the web ``\url{https://www.wikidata.org/wiki/Qxxxx}'' by replacing ``Qxxxx'' with the specific Wikidata ID. An example is shown in the following:
\begin{table}[h]
    \centering
    \scalebox{0.8}{
    \begin{tabular}{c c p{4cm}}
    \toprule
    \textbf{ID} & \textbf{Name} & \textbf{Description} \\ \midrule
         Q68 & computer & general-purpose device for performing arithmetic or logical operations  \\ \bottomrule
    \end{tabular}
    % \caption{Caption}
    \label{tab:anno1}
    }
\end{table}

If an instance is a concept, you should give the correct label. You should give the wrong label in these circumstances:
\begin{enumerate}
    \item The instance is more like an entity than a concept, such as ``Voice over Internet Protocol'' (a network protocol).
    \item The description and name of the instance are ``None''.
    \item The instance is used for the  website's construction and is meaningless, such as ``Wikimedia list article'' and ``Wikimedia disambiguation page''.
    \item Other cases that are difficult to judge.
\end{enumerate}

\paragraph{Annotation results.} We ask human annotators to annotate all concepts in KACC-L. The result of an instance is obtained if two annotators reach the agreement. If not, a third annotator is asked to label the instance. As a result, 482 concepts are removed among 15,642 concepts.

\subsection{Annotations for Multi-hop Hierarchical Triples}
\label{sec:high-anno-appendix}

In this task, we extract multi-hop ``\texttt{instanceOf}'' and ``\texttt{subclassOf}'' transitivity links from different train sets and ask annotators to label the position where the hierarchical transitivity holds.

\paragraph{Task Guidelines.} This task aims to annotate the transitivity link of concepts. For an example we provided, you need to determine whether semantic drift exists in this link and label the final position that the transitivity holds. Here is some preliminary knowledge:
\begin{itemize}
    \item \textbf{Concept.} A word for a group or a class of things, such as ``artist'', ``writer'', etc. Humans obtain concepts by abstracting commonalities from things.
    \item \textbf{Entity.} A specific person or thing, such as ``Barack Obama'', ``Mona Lisa'', etc.  
    \item \textbf{Transitivity of concepts.} An example of a transitivity link of concepts is ``$ \emph{scientist} \rightarrow \emph{researcher} \rightarrow \emph{occupation} \rightarrow \emph{human activity}$''. The transitivity link starts from an entity or a concept and follows by concepts. The transitivity of concepts assumes that the semantic of the later concept could contain the former concept. For example, the semantic of ``\emph{occupation}'' contains ``\emph{scientist}'', while ``\emph{occupation}'' is also a more general meaning concept.
    \item \textbf{Semantic drift.} Because of the annotation process of the original data source (Wikidata), we can assume almost all one-hop links are correct, such as ``$\emph{scientist} \rightarrow \emph{researcher}$'' in our example. But semantic drift occurs as the transitivity link goes deep. For example, ``\emph{scientist}'' can be subclass of ``\emph{occupation}'' while it cannot belong to ``\emph{human activity}''. However, the one-hop link ``$\emph{occupation} \rightarrow \emph{human activity}$'' still holds true.
\end{itemize}

We provide you the transitivity links of concepts with length 4. These links start from an entity or a concept and is followed by concepts. We provide you the Wikidata ID and the name of the entities and concepts. For more details, you can go to the web ``\url{https://www.wikidata.org/wiki/Qxxxx}'' by replacing ``Qxxxx'' with the specific Wikidata ID. Some examples of this task are shown in Table~\ref{tab:anno2}:

\begin{table*}[]
\centering
\scalebox{0.8}{
\begin{tabular}{l}
\toprule
\textbf{Links} \\
\midrule
(1) Q4442912 capital of Russia $\rightarrow$ Q5119 capital $\rightarrow$ Q515 city $\rightarrow$ Q702492 urban area \\
(2) Q3108101 tropical garden $\rightarrow$ Q1107656 garden $\rightarrow$ Q386724 work $\rightarrow$ Q15401930 product \\ \bottomrule
\end{tabular}
}
\caption{Annotation examples for concept transitivity.}
\label{tab:anno2}
\end{table*}

You need to label the last position that the concept transitivity holds true starting from the first entity/concept. For example (1) in Table~\ref{tab:anno2}, ``\emph{capital of Russia}'' can be regarded as a sub-concept of ``\emph{urban area}'', so the position is 4. In example (2), ``\emph{tropical garden}'' belongs to ``\emph{garden}'' while it does not belong to ``\emph{work}'', so the position can be labeled as 2.

We can assume that most one-hop links are correct, and you have no need to check the authenticity of them. For example in ``\emph{Dewey County} $\rightarrow$ \emph{county of Oklahoma}'', you do not need to check whether \emph{Dewey} is a county of \emph{Oklahoma}. However, in some specific circumstances, the one-hop link may be wrong, then you can label the case as 1, which means that only the first entitiy/concept is true.

If you cannot find out meaningful names for entities or concepts, or you meet other cases that are difficult to judge, you can label them as 0.

\paragraph{Annotation Results.} We extract 1200, 3000 and 6000 multi-hop ``\texttt{instanceOf}'' and ``\texttt{subclassOf}'' triples for KACC-S, KACC-M and KACC-L. These numbers are similar to numbers of  ``\texttt{subclassOf}'' triples in corresponding valid and test sets. We ask two annotators to annotate them, and a third annotator will be added if the two annotators do not reach an agreement. Note that our task requires to label the position, thus there are cases where all these three annotators give different labels. In these cases, we just omit these examples. If a case is labeled as 4, then we can construct both 2-hop and 3-hop triples from the link. If the case is labeled as 3, we can only obtain the 2-hop triple. The statistics of our datasets are in Table~\ref{tab:anno2-result}.

\begin{table*}[]
    \centering
    \scalebox{0.8}{
    \begin{tabular}{c c c c c c c} \toprule
    \multirow{2}{*}{\textbf{Dataset}} & \multicolumn{3}{c}{\texttt{instanceOf}} & \multicolumn{3}{c}{\texttt{subclassOf}} \\ \cline{2-7}
     & \# Extracted Links & \# 2-hop Links & \# 3-hop Links & \# Extracted Links & \# 2-hop Links & \# 3-hop Links \\  \midrule
       KACC-S & 1,200 & 1,159 & 1,137 & 1,200 & 1,170 & 1,148 \\
       KACC-M & 3,000 & 2,888 & 2,854 & 3,000 & 2,946 & 2,904 \\
       KACC-L & 6,000 & 5,723 & 5,671 & 6,000 & 5,887 & 5,806 \\ \bottomrule
    \end{tabular}}
    \caption{Statistics of annotated multi-hop triples.}
    \label{tab:anno2-result}
\end{table*}

\subsection{Statistics of Dataset Split}
\label{sec:data-appendix}
The statistics of the datasets after partition are shown in Table~\ref{tab:data-split}.

\begin{table}[h]
\begin{center}
\scalebox{0.8}{
\begin{tabular}{l c c c }
\toprule 
\textbf{Data Source} & \textbf{\# Train} & \textbf{\# Valid} & \textbf{\# Test} \\
\midrule
$\mathcal{T}_{EC}$ & 644,332 & - & - \\
\quad $\mathcal{T}_E$ & 533,209 & - & - \\
\quad $\mathcal{T}_S$ & 98,553 & - & - \\
\quad $\mathcal{T}_C$ & 12,570 & - & - \\ \midrule

$\mathcal{T}_E$ & 533,209 & 64,965 & 64,476 \\ \midrule

$\mathcal{T}_C$ & - & 1495 & 1549 \\ 
\quad $\mathcal{T}_{\texttt{sub}}$ & - & 931 & 965\\ 
\quad $\mathcal{T}_{C(\texttt{logic})}$ & - & 366 & 369\\
\quad $\mathcal{T}_{C(\texttt{ins})}$ & - & 198 & 215 \\ \midrule

$\mathcal{T}_{\texttt{ins}}$ & - & 12,679 & 12,523\\
\quad $\mathcal{T}_S$ & - & 12,481 & 12,308\\
\quad $\mathcal{T}_{C(\texttt{ins})}$ & - & 198 & 215 \\ \midrule

$\mathcal{T}_{\texttt{sub}}$ & - & 931 & 965\\ \midrule

$\mathcal{T}_{C(\texttt{logic})}$ & - & 366 & 369 \\ \midrule
$\mathcal{T}_{\texttt{M-Ins}}$ & - & 2,871 & 2,871\\
$\mathcal{T}_{\texttt{M-Sub}}$ & - & 2,925 & 2,925\\ 
\bottomrule
\end{tabular}}
\end{center}
\caption{\label{tab:data-split} Statistics of data split for KACC-M.}
\end{table}

\subsection{Additional Domain Plot for \name}
\label{sec:appendix-domain}
\begin{figure}[htbp!]
\vspace{-1em}
\centering
\subfigure{
\begin{minipage}[t]{0.505\linewidth}
\centering
\includegraphics[width=\linewidth]{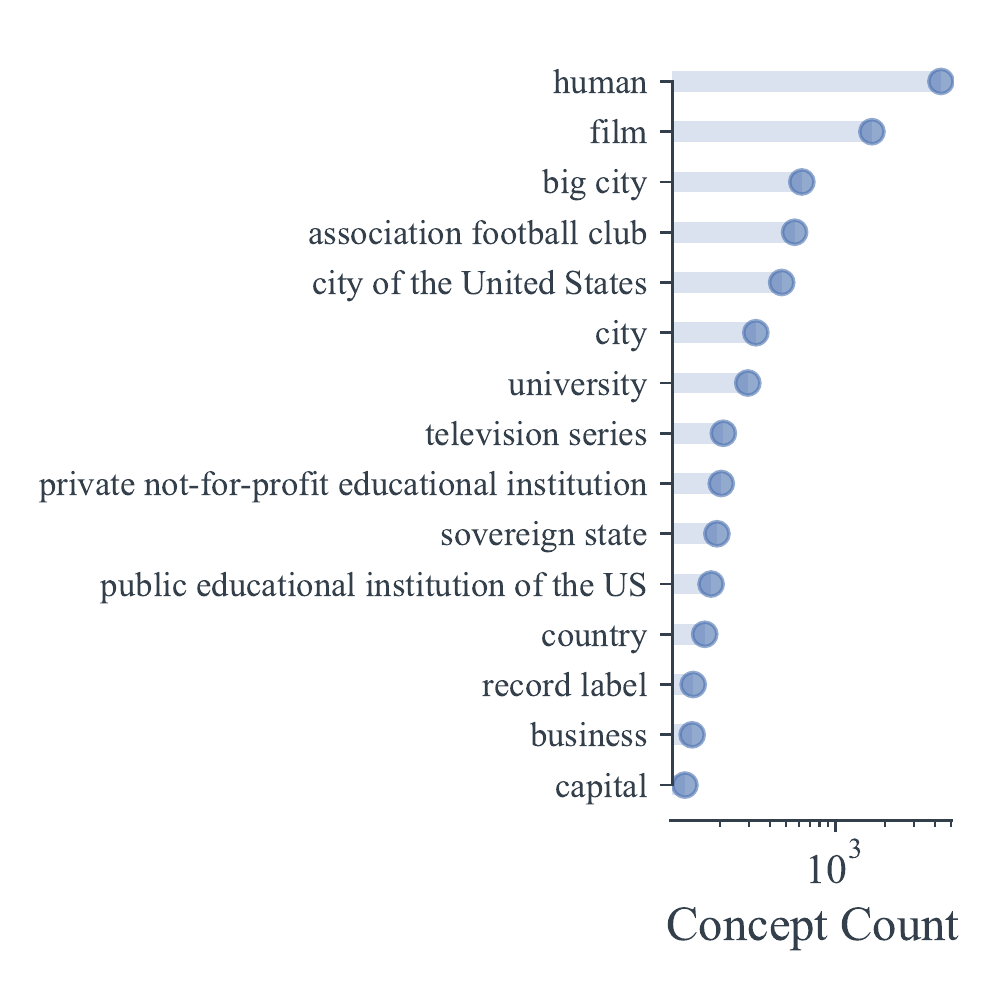}
\end{minipage}%
}
\subfigure{
\begin{minipage}[t]{0.445\linewidth}
\centering
\includegraphics[width=\linewidth]{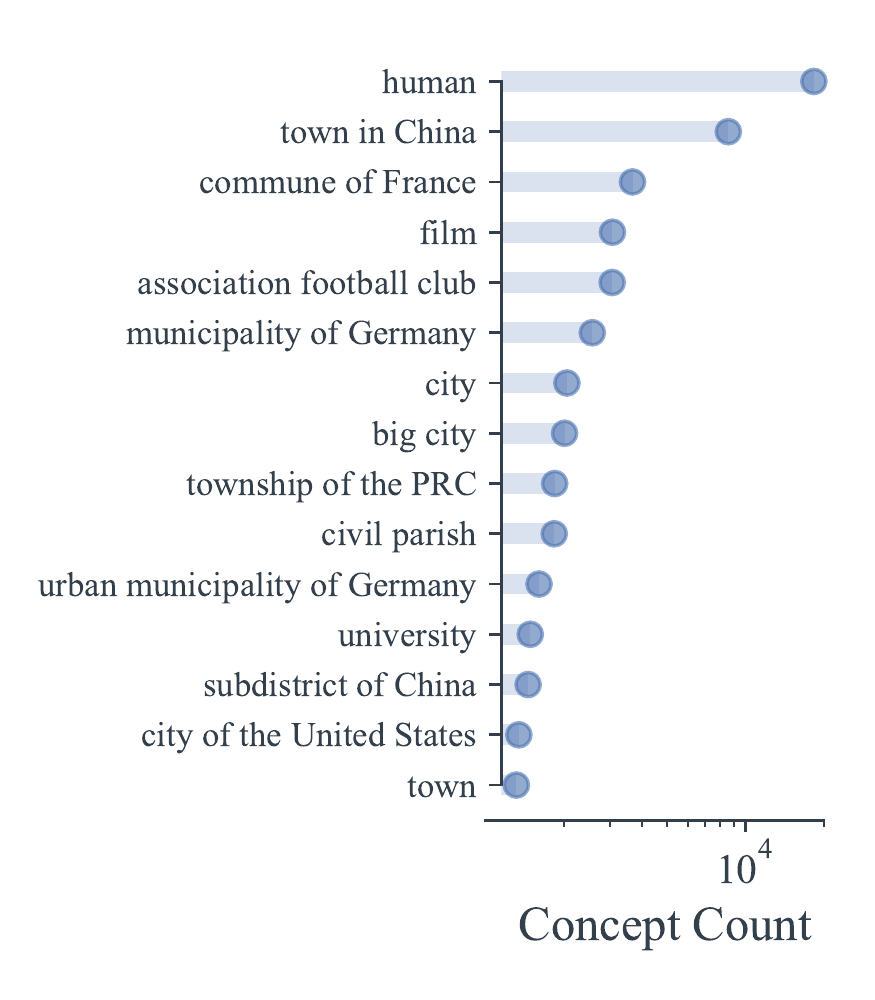}
\end{minipage}
}%
\centering
\vspace{-1em}
\caption{Top 15 most frequent bottom concepts of \name-S (left) and \name-M (right).}
\label{fig:domain-2}
\end{figure}

We plot the domains of our KACC-S and KACC-M in Figure~\ref{fig:domain-2}. Domains of KACC-S and KACC-M are similar while KACC-M has more fine-grained concepts, such as ``\emph{town in China}'' and ``\emph{commune of France}''.

\subsection{Hyperparameter Settings}
\label{sec:param-space-appendix}
In this section, we present our hyperparameter selection methods in detail. We run 30 quasi-random hyperparameter search trails on predefined hyperparameter spaces for different baselines (see Table ~\ref{tab:params-ke} to Table~\ref{tab:params-joie}). Because we use different implementations, thus hyperparameter spaces are different for different methods.

We run each trail for 100 epochs and save the checkpoint every 20 epochs (150 saved checkpoints for one model in total). Since our benchmark contains multiple tasks, for each task, we use the corresponding valid set to choose the best checkpoint based on the MRR metric, and then we test the selected checkpoint on the test set and compute final metrics. 

\newcommand\mytab{\;\;\;\;\;}
\begin{table}[h]
\begin{center}
\scalebox{0.7}{
\begin{tabular}{l  c}
\toprule 
\textbf{Hyperparameter} & \textbf{Search Range}   \\ \midrule
Training &\\
\mytab Scheme (Complex, Distmult, TuckER) & 1vsAll \\
\mytab Scheme (TransE)& Negative Sampling\\
\mytab No. subject samples (TransE) & [1, 1000], log scale\\
\mytab No. object samples (TransE) & [1, 1000], log scale\\
\mytab Batch size & \{128, 256, 512\} \\
\mytab Loss type & \{CE\} \\ \midrule
Optimizer & \{Adam, Adagrad\}\\
\mytab Learning rate & [0.001, 0.1], log scale \\
\mytab Learning rate scheduler's patience & [0, 10] \\ \midrule
Embedding &\\ 
\mytab Dimension & \{100, 200, 300\}\\
\mytab Initialization & \\
\mytab\mytab Std. deviation (Normal) & $[10^{-4}, 1.0]$, log scale \\
\mytab\mytab Interval (Uniform) & [-1.0, 1.0] \\
\mytab Regularization & \{None, $L_3$, $L_2$, $L_1$ \}\\
\mytab\mytab Entity emb.weight & $[10^{-20}, 10^{-1}]$, log scale \\
\mytab\mytab Relation emb.weight & 
$[10^{-20}, 10^{-1}]$, log scale \\
\mytab\mytab Frequency weighting &\{True, False\}\\
\mytab Dropout \\
\mytab\mytab Entity emb.dropout & $[0, 0.5]$\\
\mytab\mytab Relation emb.dropout & $[0, 0.5]$ \\
\bottomrule
\end{tabular}}
\end{center}
\caption{\label{tab:params-ke} Hyperparameter space of quasi-random search for TransE, DistMult, ComplEx, TuckER.}
\end{table}

\begin{table}[h]
\begin{center}
\scalebox{0.7}{
\begin{tabular}{l  c}
\toprule 
\textbf{Hyperparameter} & \textbf{Search Range}   \\ \midrule
Training &\\
\mytab Scheme & Negative Sampling\\
\mytab No. negative samples & \{1, 10, 20, 50\}\\
\mytab Batch size & \{128, 256, 512\} \\
\mytab Loss type & \{F2, N3\} \\ \midrule
Optimizer & \{Adam, Adagrad\}\\
\mytab Learning rate & \{0.001, 0.005, 0.01, 0.05, 0.1\} \\
Embedding &\\ 
\mytab Dimension & \{100, 200, 300\}\\
\bottomrule
\end{tabular}}
\end{center}
\caption{\label{tab:params-atth} Hyperparameter space of quasi-random search for AttH.}
\end{table}

\begin{table}[h]
\begin{center}
\scalebox{0.7}{
\begin{tabular}{l  c}
\toprule 
\textbf{Hyperparameter} & \textbf{Search Range}   \\ \midrule
Training &\\
\mytab Backend & \{TransE\} \\
\mytab Transition method & \{CG, CT\} \\
\mytab Scheme & Negative Sampling \\
\mytab Batch size & \{128, 256, 512\} \\
\mytab $a_1$, $a_2$ & \{1.0, 2.5\} \\
\mytab $m_1$, $m_2$ & \{0.5, 1.0\} \\
\midrule
Optimizer & \\
\mytab Learning rate & \{0.0005, 0.001, 0.01\} \\
Embedding &\\ 
\mytab Entity dimension & \{50, 100, 200, 300\}\\
\mytab Concept dimension & \{50, 100, 200, 300\}\\
\bottomrule
\end{tabular}}
\end{center}
\caption{\label{tab:params-joie} Hyperparameter space of quasi-random search for JOIE.}
\end{table}

\subsection{Runtime Environment}
All experiments are conducted on a server with the following environment.

\begin{itemize}
	\item Operating System: Ubuntu 18.04.3 LTS
	\item CPU: Intel(R) Xeon(R) Gold 5218 CPU @ 2.30GHz
	\item GPU: GeForce RTX 2080 Ti
\end{itemize}